\documentclass[11pt,letterpaper]{article}
\usepackage{tdal_preprint}
\usepackage{graphicx}
\usepackage{amsmath}
\usepackage{booktabs}
\usepackage{longtable,array}
\usepackage{placeins}
\usepackage{needspace}
\usepackage{algorithm}
\usepackage{algpseudocode}
\usepackage[authoryear,round]{natbib}
\setcitestyle{citesep={;},aysep={,},yysep={;}}
\usepackage[hidelinks,unicode]{hyperref}
\usepackage{url}
\title{Beyond Fixed Acquisition Functions:\\
Temporal Difference Active Learning\\
for Perturbation Discovery}
\author{
Xudong Zhang$^{1,2,*}$ \quad Jianzhe Li$^{1,2,*}$ \quad
Yitong Tang$^{1}$ \quad Peijie Zhou$^{1,2,3,\dagger}$ \\
{\small\normalfont $^{1}$Center for Machine Learning Research, Peking University, Beijing, China} \\
{\small\normalfont $^{2}$AI for Science Institute (AISI), Beijing, China} \\
{\small\normalfont $^{3}$Center for Quantitative Biology, Peking University, Beijing, China} \\
{\normalfont\texttt{xudongzhang25@stu.pku.edu.cn}} \\
{\small\normalfont $^{*}$Equal contribution.\quad $^{\dagger}$Corresponding author.}
}
\hypersetup{pdfauthor={Xudong Zhang, Jianzhe Li, Yitong Tang, Peijie Zhou}}

\date{}
\hypersetup{
  pdftitle={Beyond Fixed Acquisition Functions: Temporal Difference Active Learning for Perturbation Discovery},
  pdfsubject={Active learning for genetic perturbation discovery},
  pdfkeywords={active learning, experimental design, dynamic acquisition function, temporal difference learning, perturbation discovery}
}

\begin{document}
\maketitle
\begin{abstract}
Genetic and drug perturbation screens are important tools for
elucidating disease mechanisms and validating therapeutic targets, but
experimental costs limit the number of candidates that can be tested.
Existing perturbation screening methods typically update the surrogate model
with feedback while keeping the acquisition function fixed, limiting their
ability to adapt selection to different response landscapes.
Experimental feedback reveals current hits and informs subsequent selection
by updating candidate predictions, providing a training signal for learning
how to select experiments.
We formulate budgeted perturbation discovery as a sequential decision
problem and propose Temporal Difference Active Learning (TDAL), which learns
an acquisition function online through temporal difference learning from a
single screening trajectory, accounting for current hits and future discoveries.
TDAL uses the Hellinger distance between hit probabilities for unmeasured
candidates before and after an experiment to weight successor value, and
reuses measured outcomes through leave-batch-out (LBO) to construct virtual
transitions and alleviate training data scarcity.
Across eight public screens spanning single, double, and triple perturbations
under varied experimental configurations, TDAL achieves an average rank of
1.63 among eleven methods and ranks first on four tasks, demonstrating its
adaptability across response landscapes.
\end{abstract}

\section{Introduction}
\label{sec:introduction}

Large-scale perturbational screens deepen the understanding of
disease mechanisms and help validate therapeutic targets
\citep{b01_schmidt2022crispr,b02_zhuang2019genome,b03_sanchez2021genome}.
The human genome comprises nearly 20,000 single-gene candidates, roughly
200 million pairs, and one trillion triples in terms of perturbation combinations.
The costs of testing each perturbation and measuring its phenotype make
exhaustive screening infeasible
\citep{p02_mehrjou2022genedisco,p07_pacchiano2023neural}.
Closed-loop systems have shown that model-guided experiment selection can
improve biological discovery under limited resources
\citep{b26_xu2026lumilab,b27_demeo2025drugreflector}.
It is thus important to develop strategies that maximize cumulative discoveries
within a fixed experimental budget.

\begin{figure}[t]
    \centering
    \includegraphics[width=\linewidth]{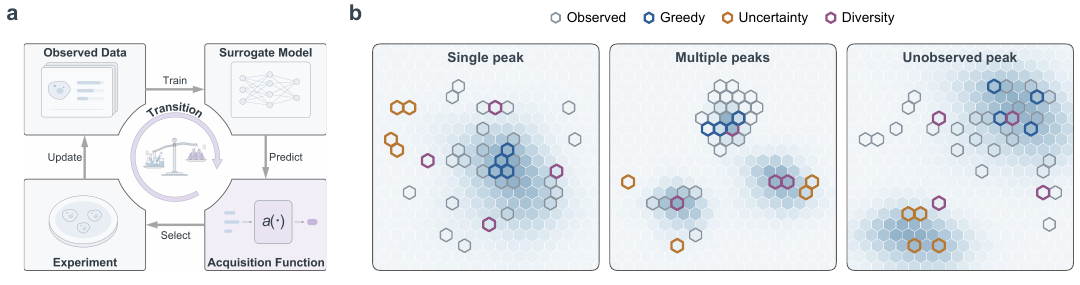}
    \caption{\textbf{a,} Active learning fits a surrogate model, selects
    experiments with an acquisition function, and updates the model with
    new observations (gray arrows). TDAL learns a dynamic acquisition
    function from accumulated transitions (purple path).
    \textbf{b,} A surrogate model is fitted to observations. Fixed
    acquisition functions (Greedy, Uncertainty, and Diversity) use its predictions
    to select the next batch, yielding different selections across illustrative
    landscapes. Hexagon fills show responses. Gray and colored outlines mark
    observed candidates and next batch selections, respectively.}
    \label{fig:intro-overview}
\end{figure}

Active learning addresses this challenge by iteratively fitting a surrogate
model to measured outcomes and using an acquisition function to select
subsequent experiments (Figure~\ref{fig:intro-overview}a).
GeneDisco evaluates active learning methods for perturbation screening and reports substantial variation in their performance across experimental settings \citep{p02_mehrjou2022genedisco}.
Existing methods employ different selection criteria to guide
perturbation discovery.
OAE \citep{p07_pacchiano2023neural} seeks near-optimal perturbations,
DiscoBAX \citep{p08_lyle2023discobax} considers response strength
and mechanistic diversity, and Probability-of-Hit (PoH)
\citep{p01_rubbi2026many} prioritizes perturbations that exceed
a predefined phenotypic threshold.
Recent approaches also incorporate language model agents or
biological priors into experiment selection
\citep{p12_roohani2025biodiscoveryagent,p37_hao2025perturboagent,p11_li2026biobo}.

Indeed, different response landscapes favor different selection strategies
(Figure~\ref{fig:intro-overview}b).
In conventional active learning, experimental feedback updates the surrogate model and changes subsequent selections, while the acquisition rule itself typically remains predefined. The same feedback can also inform how experiments should be selected:
each experiment yields immediate discoveries and updates predictions for
the remaining candidates, influencing future discovery opportunities.
This motivates learning an acquisition function that accounts for both
current and future discoveries from experimental feedback.
We therefore formulate screening as a sequential decision problem and use
experimental transitions to learn a dynamic acquisition function online
(Figure~\ref{fig:intro-overview}a).
However, unlike reinforcement learning with repeated episodes, a screening
campaign follows a single trajectory that cannot be reset, with each
candidate queried at most once and only one real transition generated per round.
Learning an acquisition function from such limited feedback presents
a particular challenge.

Here we introduce Temporal Difference Active Learning (TDAL),
which learns a dynamic acquisition function online from experimental feedback
to maximize cumulative discoveries.
Building on surrogate hit probabilities, TDAL uses temporal difference
learning to incorporate observed discoveries and feedback-weighted
successor value into experiment selection.
It weights successor value according to changes in predicted hit
probabilities for the remaining candidates, reflecting how experimental
feedback updates their predicted outcomes.
To alleviate the scarcity of real transitions, TDAL constructs additional
training transitions from previously measured outcomes through
leave-batch-out (LBO).
Our main contributions are threefold.

\begin{itemize}
    \item We formulate perturbation discovery as a sequential
    decision problem and learn a dynamic acquisition function from
    experimental feedback to adapt to different response landscapes.

    \item We introduce TDAL, which learns online from a single screening
    trajectory through temporal difference learning. It uses the Hellinger
    distance to weight successor value and LBO to construct additional
    training transitions, alleviating training data scarcity.

    \item We evaluate TDAL across eight screens spanning single,
    double, and triple perturbations. TDAL ranks first on four tasks and
    achieves an average rank of 1.63 among eleven methods.
\end{itemize}

\label{sec:introduction-end}

\section{Related Work}
\label{sec:related-work}

\subsection{Active Learning and Acquisition Function Learning}

Active learning alternates between sample selection and model
updates to improve predictions with fewer labels.
Methods include BALD, which uses information gain
\citep{c05_houlsby2011bald}, CoreSet, which emphasizes geometric coverage
\citep{c06_sener2018coreset}, and BADGE, which combines uncertainty and
diversity \citep{c08_ash2020badge}.
For discovery, active search \citep{p03_garnett2012bayesian}, ENS
\citep{p04_jiang2017efficient}, and its batch extension
\citep{p05_jiang2018efficient} evaluate future hits within the budget.
IDS \citep{p38_xu2022adaptive} balances instantaneous regret and information gain
to increase cumulative response.
These methods adapt sampling to observations but retain predefined acquisition
functions.

Other approaches learn acquisition functions from data.
Learning Active Learning predicts expected error reduction
\citep{c12_konyushkova2017learning}.
MetaBO trains acquisition functions through reinforcement learning across
tasks \citep{c13_volpp2020metabo}.
Amortized active search trains a neural network to imitate ENS selections
\citep{p06_nguyen2024amortized}.
In molecular discovery, GLARE updates its selection policy online using
measured activity labels \citep{p47_chen2025glare}, while
\citet{p39_zhang2026policy} learn a policy to identify the optimal molecule.
SOFisher uses Double DQN to select spatial-omics fields of view
\citep{p40_li2026sofisher}.
Our work focuses on perturbation screens, where TDAL learns a dynamic
acquisition function online from experimental feedback.

\subsection{Active Learning for Perturbation Discovery}

Active learning guides the discovery of perturbations with target
phenotypes under a limited experimental budget.
GeneDisco provides a unified benchmark for gene perturbations
\citep{p02_mehrjou2022genedisco}.
OAE seeks near-optimal perturbations with high responses
\citep{p07_pacchiano2023neural}, DiscoBAX balances response strength and
mechanistic diversity \citep{p08_lyle2023discobax}, and PoH selects candidates by their
posterior probability of exceeding a predefined phenotypic threshold
\citep{p01_rubbi2026many}.
RECOVER and NAIAD target drug and genetic combination discovery, respectively
\citep{p28_bertin2023recover,p14_qin2025active}, while BATCHIE uses
information-driven batch design to improve predictions of drug combination responses
\citep{p29_tosh2025batchie}.
These methods update surrogate models from observations while typically
keeping acquisition functions fixed.

Biological priors and agents guide selection.
IterPert improves predictions with multimodal priors
\citep{p13_huang2023sequential}, while BioBO combines gene representations
and pathway information \citep{p11_li2026biobo}.
\citet{p15_soper2025knowledge} and DeepAL \citep{p48_zhu2026deepal} combine
knowledge graphs with Bayesian matrix factorization and representation
learning, respectively, for double gene perturbation discovery.
BioDiscoveryAgent and PerTurboAgent design perturbations using language
models, experimental history, and tools
\citep{p12_roohani2025biodiscoveryagent,p37_hao2025perturboagent}.
AIDO.Builder updates models from feedback and reuses experience through
memory \citep{guo2026aidobuilder}.
\citet{p25_gupta2025llms} found the tested agents insensitive to permuted
feedback, whereas \citet{wainrib2026feedback} reported discovery gains
from feedback beyond prior knowledge.
We construct transitions from feedback to learn a dynamic acquisition
function that adapts to different response landscapes
(Appendix~\ref{app:method-comparison}).

\label{sec:related-work-end}

\section{Problem Formulation}
\label{sec:problem-formulation}

We consider a fixed candidate pool and aim to maximize expected cumulative
discoveries of perturbations that induce desired phenotypic responses within
a fixed experimental budget.
We formulate this task as a sequential decision problem and define an
objective combining immediate discoveries with successor value weighted
according to experimental feedback (notation in Appendix~\ref{app:notation}).

\subsection{Sequential Perturbation Discovery}
\label{sec:sequential-perturbation-discovery}

Let $\mathcal X$ be a finite pool of gene or gene combination
perturbations. Testing a candidate $x\in\mathcal X$ yields a scalar
phenotype $e(x)$. We align effect directions across assays so that larger $e(x)$
indicates closer agreement with the target phenotype.
For a threshold $\tau$ fixed before screening, the hit indicator is
$z(x)=\mathbf 1\{e(x)\ge\tau\}$ \citep{p01_rubbi2026many}.
Let $D_0$ contain the $n_0$ candidate--phenotype pairs $(x,e(x))$
observed before sequential selection begins. The phenotypes of initially
unmeasured candidates $x\in U_0$ are revealed only when queried.
The subsequent screening process consists of $T$ rounds with
batch size $B$, subject to the experimental budget.

\begin{figure}[t]
    \centering
    \includegraphics[width=\linewidth]{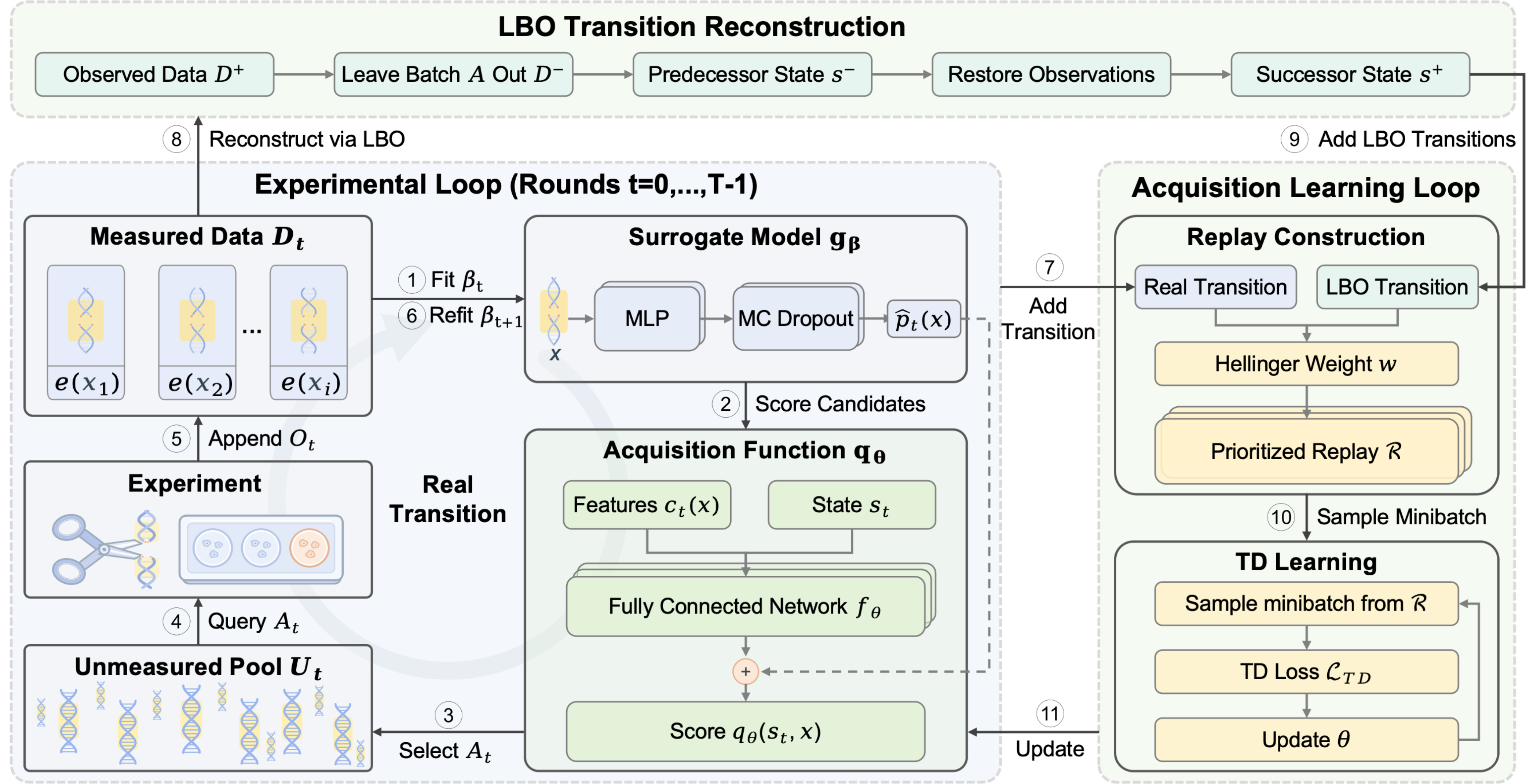}
    \caption{\textbf{TDAL framework.}
    The surrogate is fitted to measurements, and the acquisition function sums
    hit probabilities and network outputs to select the next batch.
    TDAL incorporates new observations, refits the surrogate, and constructs
    real transitions. These enter replay alongside virtual LBO transitions
    reconstructed from measurements, training the acquisition function through
    TD learning with Hellinger distance weighting successor value according
    to prediction changes.}
    \label{fig:tdal-overview}
\end{figure}

In the experimental loop shown in Figure~\ref{fig:tdal-overview}, let $A_t$
denote the selected batch at round $t\in\{0,\ldots,T-1\}$ and
$O_t=\{(x,e(x)):x\in A_t\}$ its observed outcomes.
Before selecting $A_t$, the available history is
$H_t=(D_0,(A_j,O_j)_{0\le j<t})$.
The measured dataset and unmeasured candidate pool are
\begin{equation}
\label{eq:measured-and-unmeasured}
D_t=D_0\cup\bigcup\nolimits_{0\le j<t}O_j,
\qquad
U_t=\mathcal X\setminus\{x:(x,e(x))\in D_t\}.
\end{equation}
A surrogate model trained on $D_t$ estimates hit probabilities
$\hat{p}_t(x)$ for $x\in U_t$.
The acquisition function selects $A_t\subseteq U_t$
using the available history and surrogate predictions.
Experimentally testing the selected batch yields $O_t$ and a reward equal
to the number of discoveries in round $t$:
\(
R_t=\sum\nolimits_{x\in A_t}z(x).
\)
Appending $(A_t,O_t)$ to the history gives $H_{t+1}$, 
with $D_{t+1}=D_t\cup O_t$ and $U_{t+1}=U_t\setminus A_t$.
Each round thus produces a transition $(H_t,A_t,R_t,H_{t+1})$.
After $T$ rounds, the final output consists of all verified hit candidates
recorded in $D_T$, whose cumulative discovery count is
\(
C_T=\sum\nolimits_{(x,e(x))\in D_T}z(x).
\)

% Include the approved Figure 2 near its first discussion in this partial draft.
% Its final placement will be reviewed when Section 4 is incorporated.

\subsection{Sequential Decision Objective}
\label{sec:sequential-decision-objective}

The screening goal is to maximize expected final discoveries
$\mathbb E_\pi[C_T\mid H_0]$.
Each experiment yields the current discovery count and updates the surrogate
model's predictions for unmeasured candidates, informing estimates of future
discovery value and selection in the next round.
We use the current discovery count as the immediate reward and adjust the
successor value weight according to prediction changes before and after the
experiment, defining a surrogate learning objective weighted by experimental
feedback. Let $w_t=w(H_t,A_t,H_{t+1})\in[0,1]$ denote the successor value
weight determined by the observed transition. For a selection policy $\pi$, define
\begin{equation}
\label{eq:weighted-objective}
J_w(\pi)
=
\mathbb E_\pi\!\left[
\sum\nolimits_{t=0}^{T-1}
\left(\prod\nolimits_{j=0}^{t-1}w_j\right)R_t
\,\middle|\,H_0
\right].
\end{equation}
We seek a selection policy $\pi$ that maximizes $J_w(\pi)$. When $w_t=1$
for all $t$, this is equivalent to maximizing expected final discoveries
$\mathbb E_\pi[C_T\mid H_0]$.

To maximize $J_w$, let $Q_t^{*,w}(H_t,A)$ denote the expected weighted cumulative discoveries
from round $t$ to $T-1$, when batch $A$ is selected given history
$H_t$ and subsequent selections follow the optimal policy.
This value combines current discoveries with weighted successor value
through the Bellman equation for a finite horizon:
\begin{equation}
\label{eq:weighted-bellman}
Q_t^{*,w}(H_t,A)
=
\mathbb E\!\left[
R_t+w_t
\max\nolimits_{A'\subseteq U_{t+1}}
Q_{t+1}^{*,w}(H_{t+1},A')
\,\middle|\,H_t,A_t=A
\right].
\end{equation}

At each round, the optimal selection policy $\pi_w^*$ selects the batch
maximizing $Q_t^{*,w}$ given the current history, thereby maximizing the
weighted cumulative discovery objective $J_w$:
\begin{equation}
\label{eq:optimal-policy}
\begin{aligned}
\pi_w^*(H_t)\in\arg\max\nolimits_{A\subseteq U_t}Q_t^{*,w}(H_t,A), 
J_w(\pi_w^*)=\max\nolimits_\pi J_w(\pi)
=\max\nolimits_{A\subseteq U_0}Q_0^{*,w}(H_0,A).
\end{aligned}
\end{equation}

\label{sec:problem-formulation-end}

\section{Temporal Difference Active Learning}
\label{sec:tdal}

Within this sequential decision framework, TDAL uses experimental feedback
to update a dynamic acquisition function online through TD learning,
accounting for current and future discoveries while adapting selection
to different response landscapes (Figure~\ref{fig:tdal-overview}).
See Algorithm~\ref{alg:tdal} (Appendix~\ref{app:tdal-procedure}).

\subsection{Dynamic Acquisition Function}
\label{sec:dynamic-acquisition}

The optimal value $Q_t^{*,w}(H_t,A)$ involves expectations over future
experimental feedback and is difficult to compute directly.
TDAL approximates this value using a dynamic acquisition function
learned online from experimental feedback and built on surrogate predictions.

\textbf{Surrogate model.}
In the active learning loop, the surrogate model approximates the posterior
predictive distribution of unmeasured outcomes, providing a reference for
estimating immediate discoveries. We use an MLP $g_\beta$ with MC dropout
and fit parameters $\beta_t$ to data $D_t$.
For each unmeasured candidate, we estimate $\hat{p}_t(x)$ as the proportion
of sampled predictions at or above threshold~$\tau$.

\textbf{Value decomposition.}
Given the measured data $D_t$,
$\widehat R_t(A)=\sum_{x\in A}\hat{p}_t(x)$ estimates the number of
discoveries in the unmeasured batch $A$.
Thus, the optimal batch value can be written as
\begin{equation}
\label{eq:optimal-value-decomposition}
Q_t^{*,w}(H_t,A)=\widehat R_t(A)+
\underbrace{
\mathbb E\!\left[
R_t-\widehat R_t(A)
+w_t\max\nolimits_{A'}Q_{t+1}^{*,w}(H_{t+1},A')
\,\middle|\,H_t,A_t=A
\right]
}_{F_t^*(H_t,A)}.
\end{equation}
The term $F_t^*(H_t,A)$ accounts for the expected error in this estimate
and the weighted value of subsequent discoveries. This decomposition allows
value learning to build on the surrogate's estimate of immediate discoveries,
which provides a reference grounded in measured data.

\textbf{Acquisition function parameterization.}
We approximate $F_t^*(H_t,A)$ using the summed outputs $F_\theta(s_t,A)$
of a fully connected acquisition network $f_\theta$.
The state $s_t$ summarizes screening progress, measured outcomes, and
surrogate prediction statistics for unmeasured candidates.
The candidate features $c_t(x)$ combine a fixed biological representation
with the surrogate model's predictions.
Adding each candidate's network output to its estimated hit probability
yields a dynamic acquisition function that scores candidates
based on $s_t$ and $c_t(x)$ (Appendices~\ref{app:candidate-representations}--\ref{app:features-network})
\begin{equation}
\label{eq:dynamic-acquisition-score}
q_\theta(s_t,x)=\hat{p}_t(x)+f_\theta(s_t,c_t(x)).
\end{equation}
We use an additive approximation $Q_\theta$ to make batch maximization
tractable. Selecting the $B$ unmeasured candidates with the highest scores maximizes
\begin{equation}
\label{eq:learned-batch-value}
Q_\theta(s_t,A)
=\sum\nolimits_{x\in A}q_\theta(s_t,x)
\approx Q_t^{*,w}(H_t,A).
\end{equation}
\label{sec:dynamic-acquisition-end}

\subsection{Temporal Difference Learning}
\label{sec:td-learning}

We train the acquisition network $f_\theta$ using real and LBO transitions
(Appendix~\ref{app:td-replay}).
Each transition provides an observed discovery count and a successor
state, supplying a learning signal for the acquisition function to adapt
its selection strategy to different response landscapes.

\textbf{Real transitions.}
After querying batch $A_t$, we obtain observations $O_t$ and
discovery count $R_t$. We refit the surrogate on $D_{t+1}$ and construct the
successor selection state $s_{t+1}$. With successor
value weight $w_t$, the round is stored as the training record
$\xi_t=(s_t,A_t,R_t,w_t,s_{t+1}).$

\textbf{Virtual transitions through LBO.}
Screening follows a single trajectory, which cannot be reset to
test alternative selection sequences within the same budget.
Each round therefore provides only one real batch transition.
LBO supplements these transitions by temporarily hiding measured
outcomes and then restoring them to reconstruct experimental
feedback. Let $D^+$ and $U^+$ denote the current measured dataset and
unmeasured pool. For a batch $A$ of measured candidates, hiding its
outcomes gives
$D^-=D^+\setminus\{(x,e(x)):x\in A\}.$
We refit the surrogate model on $D^-$ to construct the predecessor state $s^-$,
and construct the successor state $s^+$ from $D^+$ with the outcomes restored.
The discovery count $R$ is computed from the batch's observed outcomes.
We compute $w$ by comparing hit probability estimates before and after
restoration over $U^+$.
The resulting virtual transition is $\widetilde\xi=(s^-,A,R,w,s^+)$.
Real and virtual transitions enter the same replay buffer $\mathcal R$,
providing more training data without additional experiments.

\textbf{Feedback weights.}
For each candidate, hit and nonhit probabilities define a
Bernoulli distribution. For both real and virtual transitions, we use
Hellinger distance \citep{c15_hellinger1909neue} to compare these
distributions before and after feedback over the remaining pool.
Unchanged predictions across $U_{t+1}$ give $w_t=0$, while larger Hellinger distances assign
more weight to successor value:
\begin{equation}
\label{eq:hellinger-weight}
w_t^2=
\frac{\sum\nolimits_{x\in U_{t+1}}H_{\mathrm{Ber}}^2(\hat{p}_t(x),\hat{p}_{t+1}(x))}
{|U_{t+1}|},
\qquad
H_{\mathrm{Ber}}^2(p,q)=1-\sqrt{pq}-\sqrt{(1-p)(1-q)}.
\end{equation}

\textbf{Updating the acquisition function through TD learning.}
For a replay transition $\xi=(s,A,R,w,s')$, we follow
Double DQN. The current acquisition function selects the
successor batch $A'_\theta$, and the target network evaluates
its batch value $Q_{\bar\theta}(s',A'_\theta)$
\citep{c10_vanhasselt2016doubleq}.
Here, $\bar\theta$ denotes the target network parameters.
The current discovery count and successor value weighted by $w$ form the TD target $Y$.
Both real and LBO transitions use this target.
During each update, the stored surrogate predictions remain fixed,
and we optimize only $\theta$.
\begin{equation}
\label{eq:td-loss}
Y=R+wQ_{\bar\theta}(s',A'_\theta),
\qquad
\mathcal L_{\mathrm{TD}}(\theta)
=
\mathbb E_{\xi\sim\mathcal R}
\left[
\operatorname{Huber}\!\left(Q_\theta(s,A)-Y\right)
\right].
\end{equation}

\label{sec:td-learning-text-end}
\label{sec:td-learning-end}

\subsection{Theoretical Analysis}
\label{sec:theoretical-analysis}

We analyze weighted Bellman optimality, predictive improvement, and the final discovery gap.

\textbf{Proposition 1 (Weighted Bellman optimality).} For a finite horizon,
let $Q_t^{*,w}$ satisfy the Bellman recursion in equation~\ref{eq:weighted-bellman},
and let $\pi_w^*$ select a feasible batch maximizing $Q_t^{*,w}(H_t,A)$
at every reachable history. Then $J_w(\pi_w^*)=\max\nolimits_\pi J_w(\pi)$.
Thus, optimal batch selection maximizes the weighted discovery objective
(Appendix~\ref{app:weighted-bellman-proof}).

\textbf{Proposition 2 (Feedback and predictive improvement).}
Fix $H_t,A_t$ and $U=U_{t+1}$, with $|U|>0$. Let
$p_x=\Pr(z(x)=1\mid H_t,A_t)$ and
$q_x=\Pr(z(x)=1\mid H_t,A_t,O_t)$ denote hit probabilities
before and after feedback. Let $\Delta_{\mathrm{pred}}$ denote the expected
decrease in squared loss from $(p_x-z(x))^2$ to
$(q_x-z(x))^2$, averaged over $x\in U$ \citep{c15_gneiting2007proper}. With $w_t$ computed from
these probabilities, the expected predictive improvement conditional on
$H_t,A_t$ satisfies
\begin{equation}
\label{eq:predictive-learning-bound}
\mathbb E[w_t^4]
\le \Delta_{\mathrm{pred}}
=\frac1{|U|}\sum\nolimits_{x\in U}\mathbb E[(q_x-p_x)^2]
\le 2\mathbb E[w_t^2].
\end{equation}
This bounds predictive improvement using the Hellinger weight.
For a query $a$ revealing only $z(a)$, let
$r=\Pr(z(a)=1)\in(0,1)$,
$p_x^+=\Pr(z(x)=1\mid z(a)=1)$, and
$p_x^-=\Pr(z(x)=1\mid z(a)=0)$, suppressing the same conditioning.
The predictive improvement decomposes as
\begin{equation}
\label{eq:single-query-improvement}
\Delta_{\mathrm{pred}}(a)=\frac{r(1-r)}{|U|}
\sum\nolimits_{x\in U}(p_x^+-p_x^-)^2.
\end{equation}
Outcome uncertainty $r(1-r)$ multiplies the mean squared difference
between remaining predictions under the two outcomes.
Queries without hits can therefore inform selection
(Appendix~\ref{app:predictive-learning}).

\textbf{Proposition 3 (Final discovery gap).}
Under the same initial information and budget, let $\pi_C^*$ maximize
expected final discoveries and let $\pi_\theta$ select a feasible batch
maximizing $Q_\theta(s_t,A)$ at each round.
For feedback weights $w_t\in[0,1]$, write
$W_t=\prod\nolimits_{j=0}^{t-1}w_j$, with $W_0=1$.
Define the value error over feasible batches as
$\varepsilon_t^w(H_t)=\max\nolimits_A
|Q_\theta(s_t,A)-Q_t^{*,w}(H_t,A)|$
and the objective weighting bias as
$\Delta_C(w)=\mathbb E_{\pi_C^*}[\sum\nolimits_{t=0}^{T-1}(1-W_t)R_t\mid H_0]$.
The final discovery gap satisfies
\begin{equation}
\label{eq:final-discovery-gap}
\mathbb E_{\pi_C^*}[C_T\mid H_0]
-\mathbb E_{\pi_\theta}[C_T\mid H_0]
\le\Delta_C(w)
+2\mathbb E_{\pi_\theta}\!\left[
\sum\nolimits_{t=0}^{T-1}W_t\varepsilon_t^w(H_t)\mid H_0\right].
\end{equation}
The first term measures discovery return attenuated by weighting.
The second bounds selection losses from value estimation error.
Small terms imply near-optimal expected final discoveries.
For constant weights under a common value error bound, smaller weights
lower the error bound but raise the bias bound. Larger weights lower
the bias bound but raise the error bound
(Appendix~\ref{app:final-discovery-gap}).

\begingroup
\setlength{\textfloatsep}{10pt plus 2pt minus 2pt}
\setlength{\floatsep}{8pt plus 2pt minus 2pt}
\setlength{\intextsep}{8pt plus 2pt minus 2pt}
\section{Experiments and Results}
\label{sec:experiments}

We evaluate hit discovery performance across eight public genetic perturbation datasets
under fixed experimental budgets.
Then we examine how local landscape structure affects active discovery,
assess individual components through ablations, and analyze candidate
rankings over selection rounds.

\subsection{Experimental Setup}
\label{sec:experimental-setup}

\textbf{Data and budgets.}
We evaluate TDAL on eight public datasets recording phenotypic responses to
genetic perturbations, aiming to identify perturbations producing desired
phenotypic changes (Appendix~\ref{app:datasets-hits}). These comprise four single perturbation
tasks \citep{b01_schmidt2022crispr,b02_zhuang2019genome,b16_carnevale2022rasa2},
three double perturbation tasks
\citep{b26_fong2025multilineage,b25_simpson2023parp,b17_horlbeck2018mapping},
and one triple perturbation task \citep{b24_kuzmin2020duplicate}, with
candidate pools ranging from 11,067 to 214,642. Initial sample sizes,
batch sizes, and budgets vary across tasks (Table~\ref{tab:data-budgets}).

\textbf{Experimental protocol.}
We conduct retrospective sequential evaluation using measured phenotypes
from these genetic perturbation datasets. Candidate identities and fixed representations are
available at the start. Each method initially receives the phenotypes of
$n_0$ candidates, and the remaining outcomes are revealed as candidates
are selected in subsequent rounds.
We fix the hit threshold $\tau$ before screening to target the
top 5\% of candidates in the predefined effect direction
\citep{p02_mehrjou2022genedisco}. Selection aims to maximize final cumulative
hits $C_T$ within the budget, and recall
$\operatorname{Recall}_T=C_T/\sum_{x\in\mathcal X}z(x)$ measures discovery coverage.
Within each of five seeds (0--4), methods share the candidate pool,
initial data, budget, and hit threshold
(Appendix~\ref{app:experimental-protocol}).

\textbf{Baselines.}
We compare diverse baselines with publicly available implementations
(Appendix~\ref{app:baseline-methods}).
Perturbation discovery baselines include PoH \citep{p01_rubbi2026many},
OAE \citep{p07_pacchiano2023neural}, and DiscoBAX
\citep{p08_lyle2023discobax}. General acquisition functions include UCB
\citep{c04_srinivas2010gaussian}, Expected Improvement
\citep{c02_jones1998efficient}, and Thompson Sampling
\citep{c14_thompson1933likelihood}. Active learning baselines include BADGE
\citep{c08_ash2020badge}, CoreSet \citep{c06_sener2018coreset}, and
TopUncertain and KMeans-Embedding from GeneDisco
\citep{p02_mehrjou2022genedisco}. Random serves as a control.
For computational scalability, we evaluate DiscoBAX on four
single perturbation tasks.

\begin{figure}[H]
    \centering
    \includegraphics[width=\linewidth]{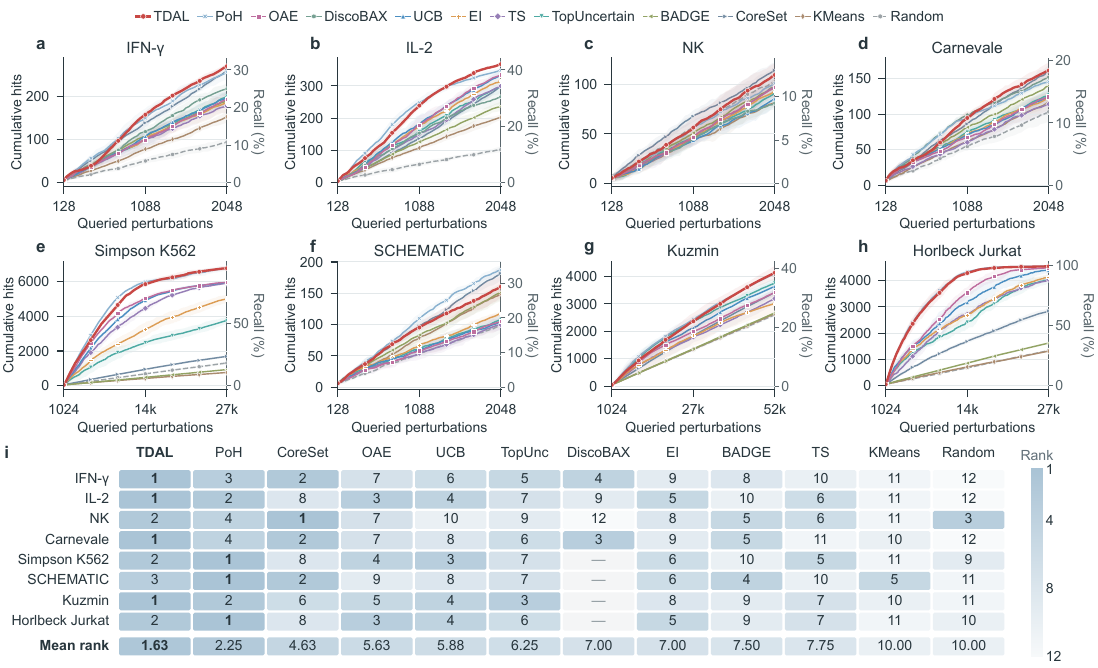}
    \caption{\textbf{Discovery performance across tasks.}
    \textbf{a--h,} Mean cumulative hits and recall with pointwise 95\%
    confidence intervals across five seeds.
    \textbf{i,} Ranks by mean final hits (darker is higher).}
    \label{fig:discovery-performance}
\end{figure}

\subsection{Budgeted Perturbation Discovery}
\label{sec:budgeted-discovery}

\textbf{Single perturbations.}
Across five seeds, TDAL achieves the highest mean final discovery counts on IFN-$\gamma$, IL-2, and Carnevale
(Figure~\ref{fig:discovery-performance}a--d).
Final cumulative hit counts are $270.6\pm6.1$, $367.4\pm3.0$,
and $161.2\pm8.6$,
corresponding to recalls of 31.00\%, 42.08\%, and 18.40\%,
respectively.
On NK, CoreSet, TDAL, and Random reach recalls of 13.01\%, 12.49\%, and
11.71\%, respectively, with small gaps between active methods and Random.
TDAL shows strong performance and adaptability across single gene perturbation landscapes.

\textbf{Combinatorial perturbations.}
TDAL ranks first on Kuzmin and third on SCHEMATIC, with final recalls of
38.59\% and 29.01\%, respectively (Figure~\ref{fig:discovery-performance}e--h).
Although ranked second on Simpson K562 and Horlbeck Jurkat, TDAL approaches
complete hit recovery with recalls of 94.40\% and 99.45\%, respectively,
within 0.2 percentage points of the best method. These results show consistently strong
performance across double and triple gene perturbation tasks.

\textbf{Stability across tasks.}
Fixed acquisition functions favor different response landscapes
\citep{p02_mehrjou2022genedisco}. Among eleven methods evaluated on all eight
main tasks, TDAL achieves an average rank of 1.63, placing first on four tasks,
second on three, and third on one (Figure~\ref{fig:discovery-performance}i).
TDAL's top three performance
on every task supports its adaptability across response landscapes
(Appendix~\ref{app:detailed-comparison}).

\begin{figure}[H]
    \centering
    \setlength{\abovecaptionskip}{5pt}
    \includegraphics[width=\linewidth]{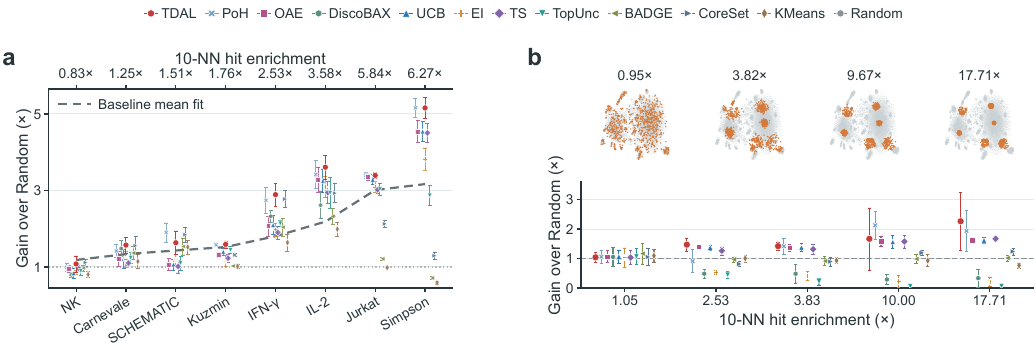}%
    \caption{\textbf{Local hit structure and discovery gains.}
    \textbf{a,} Real tasks. \textbf{b,} Noisy multicluster landscapes, with
    example hit locations. Axes show mean enrichment and discovery gains over
    Random. Points show mean gains with 95\% confidence intervals across five
    repetitions.}
    \label{fig:landscape-structure}
\end{figure}

\subsection{Landscape Structure and Active Discovery}
\label{sec:landscape-structure}

\textbf{Real landscapes.}
Active learning gains over Random vary across datasets, with several methods
underperforming Random on NK (Figure~\ref{fig:discovery-performance}).
We examine local hit structure in the fixed candidate representations.
Within an analysis sample, enrichment is the mean hit rate among each hit
candidate's 10 nearest neighbors divided by the sample's overall hit rate.
Across eight tasks, enrichment ranges from $0.83\times$ to $6.27\times$.
The gray dashed fit to mean gains across nine shared baselines
(Figure~\ref{fig:landscape-structure}a) indicates that stronger local hit
clustering is associated with greater active learning gains.

\textbf{Controlled landscapes.}
To further examine this relationship, we construct simulated landscapes
with controlled spatial structure. We keep the positions of 17,660
candidates and the set of phenotype values fixed, changing their assignment
so that hits become progressively more clustered.
With nearly random hit locations, methods perform close to Random.
Across five enrichment levels, stronger clustering yields larger gains for
TDAL and several baselines, although the benefit varies across acquisition
functions (Figure~\ref{fig:landscape-structure}b). These results indicate that
local hit structure can provide useful information for active selection.
Construction details and results are in Appendix~\ref{app:landscape}.

\subsection{Component Ablations}
\label{sec:component-ablations}

\textbf{Dynamic acquisition function.}
Surrogate only removes $f_\theta$ and selects solely by $\hat{p}_t(x)$,
updating the surrogate model while keeping the acquisition function fixed
(Appendix~\ref{app:baseline-methods}).
Its mean final discovery counts are lower than full TDAL on all four tasks, supporting learning the acquisition function alongside surrogate updates.
w/o successor retains $f_\theta$ but sets $w_t=0$, fitting
the residual $R_t-\smash[t]{\widehat R}_t(A_t)$ between observed and surrogate estimated
discoveries. Its mean final discovery count exceeds Surrogate only on all
four tasks, showing that current feedback alone can train a dynamic
acquisition function. Full TDAL adds further gains on every task,
indicating that incorporating estimated future discovery value improves
selection beyond learning from current feedback alone.

\textbf{Training and scoring design.}
Fixing $w_t=1$ lowers mean final discovery counts below both full TDAL
and w/o successor on all four tasks. Undiscounted successor value can
therefore impair selection, supporting adjustment of its weight to
experimental feedback.
Removing LBO also lowers mean final discoveries on every task, supporting
the use of additional reconstructed transitions to train $Q_\theta$ when
real experimental feedback is sparse.
w/o anchor removes the explicit $\hat{p}_t(x)$ from the score while
retaining surrogate predictions in the inputs. Its final mean decreases on
the two 60-round single perturbation tasks but increases slightly on the two
200-round tasks. With more rounds of online training, the
acquisition function maintains performance without the explicit probability baseline.
Detailed ablation settings are provided in
Appendix~\ref{app:ablations-dynamics}.

\begin{table}[H]
    \centering
    \small
    \setlength{\tabcolsep}{1.8pt}
    \begin{tabular}{@{}lcccccc@{}}
        \toprule
        Dataset & Surrogate only
        & w/o adaptive $w_t$ & w/o LBO
        & w/o successor & w/o anchor & Full TDAL \\
        \midrule
        IFN-$\gamma$ & $186.4 \pm 27.7$ & $191.8 \pm 41.1$
        & $222.2 \pm 79.8$ & $260.0 \pm 11.7$
        & $\underline{262.2 \pm 15.3}$ & $\mathbf{270.6 \pm 6.1}$ \\
        IL-2 & $253.8 \pm 15.2$ & $319.6 \pm 16.8$
        & $352.2 \pm 11.5$ & $\underline{361.4 \pm 3.4}$
        & $342.8 \pm 43.2$ & $\mathbf{367.4 \pm 3.0}$ \\
        Kuzmin & $3277.8 \pm 35.5$ & $3724.6 \pm 142.4$
        & $4096.4 \pm 208.1$ & $4086.8 \pm 89.3$
        & $\mathbf{4174.4 \pm 121.4}$ & $\underline{4141.0 \pm 194.0}$ \\
        Jurkat & $4518.2 \pm 13.3$ & $4301.2 \pm 248.3$
        & $4510.4 \pm 39.3$ & $4536.0 \pm 15.3$
        & $\mathbf{4550.2 \pm 3.8}$ & $\underline{4542.8 \pm 4.8}$ \\
        \bottomrule
    \end{tabular}
    \caption{\textbf{Core ablations on four tasks.}
    Final cumulative hits across five seeds (mean $\pm$ SD).
    Bold and underlined values mark the highest and second-highest means.}
    \label{tab:component-ablations}
\end{table}

\begin{figure}[H]
    \centering
    \includegraphics[width=\linewidth]{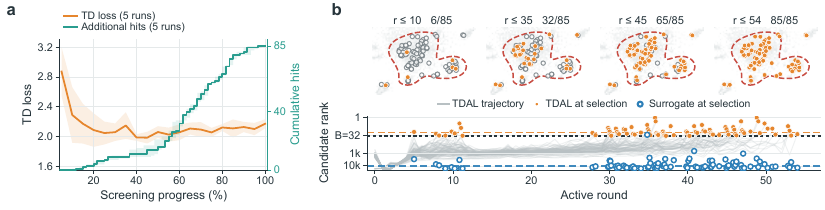}
    \caption{\textbf{Online learning and candidate reranking.}
    \textbf{a,} Training loss and cumulative discoveries over screening rounds.
    \textbf{b,} Hit discovery in a t-SNE projection of representations (top)
    and ranking trajectories (bottom), with a comparison between TDAL
    and surrogate prediction ranks.}
    \label{fig:online-feedback}
\end{figure}

\subsection{Online Feedback and Selection Paths}
\label{sec:online-feedback}
To examine how online learning changes candidate selection, we analyze
85 IL-2 hits discovered in all five TDAL runs but in none of the
Surrogate only runs. Discoveries concentrate in the second half of
screening, with the median cumulative count rising from 16 after
30 rounds to 85 at completion (Figure~\ref{fig:online-feedback}a).
Candidate ranks under TDAL progressively improve before selection
(Figure~\ref{fig:online-feedback}b).
Across the 85 candidates, their mean ranks at selection are 16.6
under TDAL and 9,780 under concurrent surrogate predictions,
showing how accumulated feedback helps prioritize hits ranked
poorly by the surrogate alone
(Appendix Figure~\ref{fig:il2-multiseed-dynamics}).

\label{sec:experiments-end}

\section{Conclusion}
\label{sec:conclusion}

We introduce TDAL to learn a dynamic acquisition function online
from a single perturbation screening trajectory.
Building on surrogate-estimated hit probabilities, TDAL uses temporal
difference learning to incorporate observed discoveries and
feedback-weighted successor value, while LBO reconstructs additional
training transitions from measured outcomes.
By learning how to select experiments from feedback accumulated
within the ongoing screen, TDAL adapts candidate selection to
both immediate discoveries and subsequent discovery opportunities
under limited experimental feedback.
Across eight public screens spanning single, double, and triple
perturbations, TDAL achieves an average rank of 1.63 among eleven
methods, demonstrating consistent discovery performance across
diverse response landscapes.

\textbf{Limitations.}
Our evaluation is retrospective, and the additive acquisition function
does not explicitly model interactions among candidates within a batch.
Feedback weighting also relies on surrogate-estimated hit probabilities,
motivating further investigation of its robustness to prediction errors.
Future work will explore batch-aware selection and alternative feedback
weighting strategies, alongside prospective validation in closed-loop
biological experiments.

\endgroup
\subsection*{Reproducibility statement}

The TDAL implementation, task configurations, and data preparation scripts
are available at \url{https://anonymous.4open.science/r/TDAL/},
with data sources and running instructions.
Appendices~\ref{app:algorithm}--\ref{app:experimental-settings} provide
implementation details, theoretical assumptions and proofs, and experimental
settings, respectively.
\subsection*{AI use statement}

The research ideas and methodological design originated from
the authors. Generative AI tools were used to assist with
code development, checking mathematical derivations and proofs,
and improving manuscript clarity and grammar.
The authors reviewed and verified all AI-assisted content
and take full responsibility for the final work.

\bibliographystyle{plainnat}
\bibliography{references}

\clearpage
\appendix
\section*{Appendix}
\section{Algorithm and Implementation Details}
\label{app:algorithm}
\renewcommand{\thetable}{\thesection\arabic{table}}
\renewcommand{\theHtable}{\thesection.\arabic{table}}
\setcounter{table}{0}

\subsection{Comparison with Related Methods}
\label{app:method-comparison}

Table~\ref{tab:method-mechanisms} compares representative methods for
perturbation screening, molecular virtual screening, and spatial omics
sampling across five mechanisms including surrogate updates, biological knowledge,
online acquisition updates, value bootstrapping, and experience
augmentation.

\begin{table}[H]
\centering
\caption{Comparison of active screening and experimental design methods.
$\checkmark$/$\times$ indicate whether a method uses each mechanism.}
\label{tab:method-mechanisms}
\footnotesize
\setlength{\tabcolsep}{1.5pt}
\renewcommand{\arraystretch}{1.28}
\begin{tabular}{@{}>{\raggedright\arraybackslash}m{\dimexpr\linewidth-312pt\relax}>{\centering\arraybackslash}m{49pt}>{\centering\arraybackslash}m{45pt}>{\centering\arraybackslash}m{42pt}>{\centering\arraybackslash}m{49pt}>{\centering\arraybackslash}m{55pt}>{\centering\arraybackslash}m{54pt}@{}}
\toprule
Method & \shortstack{Research\\task} & \shortstack{Surrogate\\updates} & \shortstack{Biological\\knowledge} & \shortstack{Online\\acquisition\\updates} & \shortstack{Value\\bootstrapping} & \shortstack{Experience\\augmentation} \\
\midrule
\shortstack[l]{GeneDisco surrogate\\baselines} & Perturbation screening & $\checkmark$ & $\times$ & $\times$ & $\times$ & $\times$ \\
\addlinespace[3pt]
OAE & Perturbation screening & $\checkmark$ & $\times$ & $\times$ & $\times$ & $\times$ \\
\addlinespace[3pt]
DiscoBAX & Perturbation screening & $\checkmark$ & $\times$ & $\times$ & $\times$ & $\times$ \\
\addlinespace[3pt]
PoH & Perturbation screening & $\checkmark$ & $\times$ & $\times$ & $\times$ & $\times$ \\
\addlinespace[3pt]
BioBO & Perturbation screening & $\checkmark$ & $\checkmark$ & $\times$ & $\times$ & $\times$ \\
\addlinespace[3pt]
BioDiscoveryAgent & Perturbation screening & $\times$ & $\checkmark$ & $\times$ & $\times$ & $\times$ \\
\addlinespace[3pt]
PerTurboAgent & Perturbation screening & $\checkmark$ & $\checkmark$ & $\times$ & $\times$ & $\times$ \\
\addlinespace[3pt]
GLARE & Molecular virtual screening & $\checkmark$ & $\times$ & $\checkmark$ & $\times$ & $\times$ \\
\addlinespace[3pt]
SOFisher & Spatial omics sampling & $\times$ & $\times$ & $\times$ & $\checkmark$ & $\times$ \\
\midrule
\textbf{TDAL} & Perturbation screening & $\checkmark$ & $\times$ & $\checkmark$ & $\checkmark$ & $\checkmark$ \\
\addlinespace[3pt]
\bottomrule
\end{tabular}
\end{table}

Surrogate updates use newly measured outcomes to fit a model that predicts
unmeasured phenotypes or activities. Biological knowledge guides selection
through pathway information, literature, or knowledge encoded in language
models. BioBO and PerTurboAgent combine such knowledge with response
modeling, whereas BioDiscoveryAgent incorporates experimental observations
into its language model context without fitting a response surrogate.
SOFisher uses observed spatial features as inputs to its sampling policy
rather than fitting a phenotype prediction model.

Online acquisition updates use feedback from the current task to update the
acquisition function's parameters. Value bootstrapping constructs
training targets from current rewards and estimated successor values.
GLARE updates its selection policy during screening but does not explicitly
bootstrap a successor value. SOFisher trains its sampling policy through
value bootstrapping before application and uses fixed policy parameters on
new tissue sections, while incorporating newly sampled observations into
the state. TDAL performs both acquisition updates and successor value
bootstrapping within the current screening task.

Experience augmentation reconstructs additional transitions from measured
outcomes to train the acquisition function. In budgeted experimental design
without replacement, the measured dataset grows while the unmeasured pool
and remaining budget shrink. A screen therefore provides a single executed
trajectory and cannot return to an earlier state to test alternative
experimental sequences within the same budget. LBO reconstructs predecessor
states in which a measured batch has not yet been observed, then restores
that batch's real outcomes to form additional transitions to a known
successor state. The same experimental outcomes can thus support batch
value learning under different information states, expanding online
training experience without additional experiments.

\Needspace{4\baselineskip}
\subsection{Notation}
\label{app:notation}

Table~\ref{tab:core-notation} summarizes the main notation used in the problem formulation and TDAL.

\begingroup
\small
\renewcommand{\arraystretch}{1.2}
\begin{longtable}{@{}>{\raggedright\arraybackslash}p{0.25\linewidth}>{\raggedright\arraybackslash}p{0.72\linewidth}@{}}
\caption{Key notation for the screening objective and TDAL.}\label{tab:core-notation}\\
\toprule Symbol & Meaning \\ \midrule
\endfirsthead
\toprule Symbol & Meaning \\ \midrule
\endhead
\bottomrule\endfoot
$\mathcal X$ & Fixed pool of candidate perturbations. \\
$e(x)$ & Phenotype of candidate $x$ after aligning the effect direction. \\
$\tau$ & Threshold defining the desired phenotypic response. \\
$z(x)$ & Indicator that candidate $x$ meets the hit threshold. \\
$D_t$ & Measured data available before selection in round $t$. \\
$U_t$ & Unmeasured candidates available in round $t$. \\
$H_t$ & History of observations and selections before round $t$. \\
$A_t$ & Batch selected for experimentation in round $t$. \\
$B$ & Number of candidates queried per round. \\
$T$ & Number of active screening rounds. \\
$R_t$ & Observed number of discoveries in batch $A_t$. \\
$C_T$ & Final verified discovery count, including initial discoveries. \\
$\pi$ & Policy that selects each batch from the available history. \\
$J_w(\pi)$ & Expected cumulative discoveries weighted by experimental feedback. \\
$Q_t^{*,w}(H_t,A)$ & Optimal weighted value of selecting batch $A$ given history $H_t$. \\
$w_t$ & Successor value weight determined by prediction changes. \\
$g_\beta$ & Surrogate model predicting perturbation phenotypes. \\
$\hat{p}_t(x)$ & Estimated hit probability of an unmeasured candidate. \\
$\widehat R_t(A)$ & Estimated immediate discovery count from surrogate hit probabilities. \\
$\phi(x)$ & Fixed biological representation of candidate $x$. \\
$s_t$ & Global summary of screening progress, observations, and predictions. \\
$c_t(x)$ & Candidate features combining biological representation and predictions. \\
$f_\theta$ & Acquisition network mapping the state and candidate features to an output. \\
$q_\theta(s_t,x)$ & Dynamic acquisition function scoring candidate $x$. \\
$Q_\theta(s_t,A)$ & Batch value estimate obtained by summing candidate scores. \\
$\xi_t$ & Transition generated by an experimental round. \\
$\widetilde\xi$ & Virtual transition reconstructed by LBO from measured outcomes. \\
$Y_i$ & TD target for transition $i$. \\
\end{longtable}
\endgroup

\subsection{Complete TDAL Algorithm}
\label{app:tdal-procedure}

Algorithm~\ref{alg:tdal} summarizes TDAL from initialization through the end of
screening. Starting from the initial observations $D_0$, TDAL fits a surrogate
model, estimates hit probabilities for candidates in the initial unmeasured
pool $U_0$, and forms the initial selection state.
It then constructs initial LBO transitions from $D_0$ to initialize the
acquisition network through TD training before the first query.

At each round, the acquisition function scores candidates in $U_t$, and
TDAL queries the $B$ candidates with the highest scores. The observed outcomes
are added to $D_t$, and the queried candidates are removed from $U_t$.
TDAL refits the surrogate model on the updated data. Before the next round,
it updates predictions over $U_{t+1}$ to construct the successor state.
Comparing hit probabilities before and after the experiment over this pool
gives the Hellinger weight. The selected batch, observed discovery count,
feedback weight, and current and successor states form a real transition.
Real and online LBO transitions enter the replay buffer, from which TD updates
train the acquisition network for subsequent selection. After $T$ rounds,
TDAL returns all verified hit candidates.

\begin{algorithm}[!htbp]
\caption{TDAL}
\label{alg:tdal}
\renewcommand{\algorithmicrequire}{\textbf{Input}}
\begin{algorithmic}[1]
\Require Candidate pool $\mathcal X$ with representations $\phi(x)$, initial data
$D_0$, rounds $T$, batch size $B$, threshold $\tau$, and LBO transition limits.
\State Initialize $U_0\leftarrow\mathcal X\setminus\{x:(x,e(x))\in D_0\}$.
\State Initialize acquisition network $f_\theta$ with a zero output layer.
\State Initialize target network $f_{\bar\theta}$ with $\bar\theta\leftarrow\theta$.
\State Initialize replay buffer $\mathcal R\leftarrow\varnothing$.
\State Fit surrogate $g_{\beta_0}$ to $D_0$ with validation early stopping.
\State Estimate $\widehat\mu_0(x)$, $\widehat\sigma_0(x)$, and $\hat{p}_0(x)$ using 50 MC dropout samples for $x\in U_0$.
\State Construct $s_0$ and candidate features $\{c_0(x):x\in U_0\}$.
\If{$|D_0|>B$}
    \State Add initial LBO transitions to $\mathcal R$ using Algorithm~\ref{alg:lbo-batches}.
    \For{20 initial TD updates}
        \State Update $f_\theta$ using the TD loss on a replay minibatch.
        \State Soft update the target parameters $\bar\theta$.
        \State Refresh the sampled transitions\textquotesingle\ replay priorities.
    \EndFor
\EndIf
\For{$t=0,\ldots,T-1$}
    \State Compute $q_\theta(s_t,x)=\hat{p}_t(x)+f_\theta(s_t,c_t(x))$ for $x\in U_t$.
    \State Select the $B$ highest scoring candidates as $A_t$.
    \State Experimentally query $A_t$ to obtain $O_t=\{(x,e(x)):x\in A_t\}$.
    \State Compute the discovery count $R_t=\sum\nolimits_{x\in A_t}z(x)$.
    \State Update measured data $D_{t+1}\leftarrow D_t\cup O_t$.
    \State Update the unmeasured pool $U_{t+1}\leftarrow U_t\setminus A_t$.
    \State Refit $g_{\beta_{t+1}}$ to $D_{t+1}$ with validation early stopping.
    \If{$t<T-1$}
        \State Estimate $\widehat\mu_{t+1}(x)$, $\widehat\sigma_{t+1}(x)$, and $\hat{p}_{t+1}(x)$ by MC dropout for $x\in U_{t+1}$.
        \State Construct $s_{t+1}$ and candidate features $\{c_{t+1}(x):x\in U_{t+1}\}$.
        \State Compute Hellinger weight $w_t$ from $\hat{p}_t(x)$ and $\hat{p}_{t+1}(x)$ over $U_{t+1}$.
    \Else
        \State Set the terminal successor state $s_T$.
        \State Set $w_t\leftarrow0$.
    \EndIf
    \State Add $\xi_t=(s_t,A_t,R_t,w_t,s_{t+1})$ and its prediction inputs to $\mathcal R$.
    \State Add online LBO transitions using Algorithm~\ref{alg:lbo-batches}.
    \For{20 online TD updates}
        \State Update $f_\theta$ using the TD loss on a replay minibatch.
        \State Soft update the target parameters $\bar\theta$.
        \State Refresh the sampled transitions\textquotesingle\ replay priorities.
    \EndFor
\EndFor
\State \Return verified hit set $\{x:(x,e(x))\in D_T,\ z(x)=1\}$.
\end{algorithmic}
\end{algorithm}

\Needspace{4\baselineskip}
\subsection{Candidate Representations}
\label{app:candidate-representations}

The mapping $\phi(x)$ encodes candidate perturbation $x$ as a vector of fixed
dimension, providing gene functional information to the surrogate and acquisition
network.

Human tasks use genome-wide CRISPR--Cas9 knockout screening data from the
Broad Institute's Project Achilles. The CERES gene dependency profiles in
DepMap 20Q4 provide 808-dimensional gene representations. Each dimension
corresponds to a cancer cell line and describes the effect of knocking out
the gene on cell growth. CERES corrects for copy number effects
\citep{b29_meyers2017ceres}. The Kuzmin yeast task uses the 500-dimensional
STRING network embeddings released with Mashup \citep{b30_cho2016mashup}.
These base gene representations remain fixed during screening.
For a single gene perturbation, $\phi(x)$ is the corresponding gene embedding.

For a candidate $x$ containing $m\in\{2,3\}$ genes with embeddings
$\mathbf v_1,\ldots,\mathbf v_m$, we define
\begin{equation}
\label{eq:combinatorial-representation}
\phi(x)=\left[
\frac{1}{m}\sum\nolimits_{i=1}^{m}\mathbf v_i,\,
\frac{2}{m(m-1)}\sum\nolimits_{i<j}|\mathbf v_i-\mathbf v_j|,\,
\frac{2}{m(m-1)}\sum\nolimits_{i<j}\mathbf v_i\odot\mathbf v_j
\right].
\end{equation}
The brackets concatenate the mean embedding, mean elementwise absolute
difference, and mean elementwise product, with $\odot$ denoting elementwise
multiplication. Averaging over all members and unordered pairs makes this
construction invariant to gene order.

\subsection{Surrogate Model and Hit Probability Estimation}
\label{app:surrogate}

\textbf{Model and training.}
The surrogate takes a fixed vector representation (embedding) $\phi(x)$ of
candidate perturbation $x$ as input and predicts the phenotype $e(x)$ after
aligning the effect direction. Its two hidden layers have widths 128 and 64.
Each applies a linear transformation followed by ReLU and dropout. A final linear layer produces a scalar output. The model is

\begin{equation}
\begin{aligned}
\mathbf h_{\mathrm{sur}}^{(1)}(x)
&=\operatorname{Dropout}_{0.2}\!\left(
\operatorname{ReLU}(W_1^{\mathrm{sur}}\phi(x)+b_1^{\mathrm{sur}})
\right)\in\mathbb R^{128},\\
\mathbf h_{\mathrm{sur}}^{(2)}(x)
&=\operatorname{Dropout}_{0.2}\!\left(
\operatorname{ReLU}(W_2^{\mathrm{sur}}\mathbf h_{\mathrm{sur}}^{(1)}(x)+b_2^{\mathrm{sur}})
\right)\in\mathbb R^{64},\\
g_\beta(\phi(x))
&=W_{\mathrm{out}}^{\mathrm{sur}}\mathbf h_{\mathrm{sur}}^{(2)}(x)
+b_{\mathrm{out}}^{\mathrm{sur}}\in\mathbb R.
\end{aligned}
\end{equation}

Here, $\operatorname{Dropout}_{0.2}$ denotes dropout with probability 0.2,
and $\beta$ contains all weights and biases of the hidden and output layers.
Predictive uncertainty is estimated through MC dropout
\citep{c09_gal2016dropout}.

At each round, the measured data $D_t$ are split into a training set
$D_t^{\mathrm{tr}}$ and a validation set $D_t^{\mathrm{val}}$ for early
stopping, and the model parameters are reinitialized.
The surrogate fits measured phenotypes by minimizing the mean squared error

\begin{equation}
\mathcal L_{\mathrm{sur}}(\beta)
=\frac{1}{|D_t^{\mathrm{tr}}|}
\sum_{(x,e(x))\in D_t^{\mathrm{tr}}}
\bigl(g_\beta(\phi(x))-e(x)\bigr)^2.
\end{equation}

Dropout is active during training and disabled when evaluating validation
mean squared error after each epoch. Let $\beta_t^{(j)}$ denote the parameters
after epoch $j$, and let $J_t$ be the number of epochs completed in the current
round. We select the final model by minimizing validation error,

\begin{equation}
\begin{aligned}
j_t^*
&=\arg\min_{1\le j\le J_t}
\frac{1}{|D_t^{\mathrm{val}}|}
\sum_{(x,e(x))\in D_t^{\mathrm{val}}}
\bigl(g_{\beta_t^{(j)}}(\phi(x))-e(x)\bigr)^2,\\
\beta_t&=\beta_t^{(j_t^*)}.
\end{aligned}
\end{equation}

\textbf{Predictive statistics.}
Dropout remains active at inference. For each unmeasured candidate $x\in U_t$,
we independently sample $K$ forward passes. Denote the $k$th stochastic
forward pass by $g_{\beta_t}^{(k)}$, giving the phenotype prediction
$\widehat e_t^{(k)}(x)=g_{\beta_t}^{(k)}(\phi(x))$.
The predictive mean, uncertainty, and hit probability are estimated by the
sample mean, standard deviation, and fraction reaching the hit threshold, respectively,

\begin{equation}
\begin{aligned}
\widehat\mu_t(x)
&=\frac{1}{K}\sum\nolimits_{k=1}^K\widehat e_t^{(k)}(x),\\
\widehat\sigma_t(x)
&=\left[\frac{1}{K}\sum\nolimits_{k=1}^K
\bigl(\widehat e_t^{(k)}(x)-\widehat\mu_t(x)\bigr)^2\right]^{1/2},\\
\hat{p}_t(x)
&=\frac{1}{K}\sum\nolimits_{k=1}^K
\mathbf 1\{\widehat e_t^{(k)}(x)\ge\tau\}.
\end{aligned}
\end{equation}

These statistics define candidate features and the selection state.
The hit probability also provides the acquisition function's base score and
is used to compute the Hellinger weight.

\subsection{State Representation and Acquisition Function Parameterization}
\label{app:features-network}

\textbf{Dynamic candidate features.}
For each unmeasured candidate $x\in U_t$, the acquisition network receives
the shared selection state $s_t$ and candidate features $c_t(x)$. These
features combine the fixed biological representation $\phi(x)$ with the
surrogate model's predicted mean, uncertainty, and hit probability. We
center the mean at the hit threshold and scale both the mean and uncertainty, giving

\begin{equation}
c_t(x)=\left[
\phi(x),\,
\frac{\widehat\mu_t(x)-\tau}{a_t},\,
\log\!\left(1+\frac{\widehat\sigma_t(x)}{a_t}\right),\,
\hat{p}_t(x)
\right].
\end{equation}

Here, $\widehat\mu_t(x)$ and $\widehat\sigma_t(x)$ are the predictive mean and
standard deviation defined in Appendix~\ref{app:surrogate}. The hit
probability $\hat{p}_t(x)$ is included without rescaling.

The normalization denominator $a_t$ is computed from the phenotypes already
measured in $D_t$. It is the larger of their interquartile range (75th minus
25th percentile) and population standard deviation, with a minimum of
$10^{-6}$ to prevent division by zero. The same denominator is used for all
candidates scored at that round. Their predicted means and uncertainties
remain specific to each candidate. We recompute $a_t$ when the measured data change.

\textbf{Selection state.}
The selection state $s_t$ is a vector summary of the current screening
history $H_t$ that provides global information for the dynamic acquisition
function. Shared by all candidates, it combines screening progress,
measured outcomes, and surrogate prediction statistics over the unmeasured
pool $U_t$. Its 15 components follow the network input order
\begin{equation}
\label{eq:selection-state}
\begin{aligned}
s_t=\Big[&
\frac{|D_t|}{M},\,
\frac{M-|D_t|}{M},\,
\frac{\lceil\min\{M-|D_t|,|U_t|\}/B\rceil}{\lceil M/B\rceil},\,
\frac{|U_t|}{|\mathcal X|},\,
\frac{\min\{B,M-|D_t|,|U_t|\}}{B},\\
&\tau,\,\max\nolimits_{(x,e(x))\in D_t}e(x),\,a_t,\,
\operatorname{Mean}_{U_t}(\hat{p}_t),\,
\operatorname{SD}_{U_t}(\hat{p}_t),\,
\operatorname{P}_{90,U_t}(\hat{p}_t),\\
&\operatorname{Mean}_{U_t}\!\left(\frac{\widehat\mu_t-\tau}{a_t}\right),\,
\operatorname{SD}_{U_t}\!\left(\frac{\widehat\mu_t-\tau}{a_t}\right),\\
&\operatorname{Mean}_{U_t}\!\left(\frac{\widehat\sigma_t}{a_t}\right),\,
\operatorname{P}_{90,U_t}\!\left(\frac{\widehat\sigma_t}{a_t}\right)
\Big]\in\mathbb R^{15}.
\end{aligned}
\end{equation}
Here, $\operatorname{Mean}_{U_t}$, $\operatorname{SD}_{U_t}$, and
$\operatorname{P}_{90,U_t}$ denote the mean, population standard deviation,
and 90th percentile over candidates $x\in U_t$.
The first five components describe the measured budget fraction, remaining
budget fraction, normalized remaining rounds, remaining pool fraction, and
query batch size relative to $B$. These ratios are clipped to $[0,1]$.
The batch size ratio is one for the full batches used here.

The hit threshold $\tau$, largest measured phenotype
$\max\nolimits_{(x,e(x))\in D_t}e(x)$, and normalization denominator $a_t$
retain the phenotype scale after aligning the effect direction.
Statistics of the hit probability $\hat{p}_t(x)$, normalized predicted effect
$(\widehat\mu_t(x)-\tau)/a_t$, and scaled uncertainty
$\widehat\sigma_t(x)/a_t$ summarize predictions over $x\in U_t$.
The uncertainty statistics use the scaled standard deviation.

\textbf{Acquisition network.}
The acquisition network $f_\theta$ is the trainable component of
$q_\theta=\hat{p}_t+f_\theta$.
For each unmeasured candidate $x\in U_t$, its encoding and network layers are
\begin{equation}
\label{eq:acquisition-network}
\begin{aligned}
u_\theta(x)
&=\operatorname{ReLU}\!\left(
W_{\mathrm{enc}}\frac{\phi(x)}{\max\{\|\phi(x)\|_2,10^{-8}\}}
+b_{\mathrm{enc}}\right)\in\mathbb R^{64},\\
\mathbf h_t^{(0)}(x)
&=\left[s_t,\,u_\theta(x),\,
\frac{\widehat\mu_t(x)-\tau}{a_t},\,
\log\!\left(1+\frac{\widehat\sigma_t(x)}{a_t}\right),\,
\hat{p}_t(x)\right]\in\mathbb R^{82},\\
\mathbf h_t^{(k)}(x)
&=\operatorname{ReLU}\!\left(W_k^{\mathrm{acq}}\mathbf h_t^{(k-1)}(x)
+b_k^{\mathrm{acq}}\right)\in\mathbb R^{128},\quad k=1,2,\\
f_\theta(s_t,c_t(x))
&=W_{\mathrm{out}}^{\mathrm{acq}}\mathbf h_t^{(2)}(x)
+b_{\mathrm{out}}^{\mathrm{acq}}\in\mathbb R.
\end{aligned}
\end{equation}
Here, $u_\theta(x)$ encodes the normalized fixed representation. The input
$\mathbf h_t^{(0)}(x)$ concatenates the 15 state features, 64 encoding
features, and three surrogate prediction features from $c_t(x)$.

The parameters $\theta$ include the weights and biases of the projection,
two hidden layers, and output layer. The static embeddings remain fixed,
while the projection is learned with the acquisition network. The output layer's
weights and bias are initialized to zero, so that $f_\theta=0$ and
$q_\theta=\hat{p}_t$ before initial TD training.

\subsection{TD Learning and Experience Replay}
\label{app:td-replay}

\textbf{LBO transition construction.}
Initial LBO hides batches from $D_0$ to train the acquisition network before
the first query. Online LBO draws hidden batches from candidates measured
after initialization, so a batch can combine observations from different
rounds. Algorithm~\ref{alg:lbo-batches} gives the construction for both stages.

The current measured dataset $D^+$ and unmeasured pool $U^+$ define the
successor state $s^+$. Each hidden batch produces a different predecessor
state, but restoring its outcomes returns to this same successor state.
For a nonterminal successor, let
$\hat{p}^+(x)$ denote the MC dropout hit probability estimated by the
surrogate trained on $D^+$, for $x\in U^+$. For initial LBO,
the successor dataset is $D^+=D_0$, giving $\hat{p}^+(x)=\hat{p}_0(x)$.
For online LBO after round $t$, it is $D^+=D_{t+1}$, giving
$\hat{p}^+(x)=\hat{p}_{t+1}(x)$.

For each hidden batch $A$, removing its outcomes from $D^+$ gives the
predecessor dataset $D^-$, and returning its candidates to the unmeasured
pool gives $U^-$. We fit the predecessor surrogate $g_{\beta^-}$ to
$D^-$ with validation early stopping.
Its MC dropout predictions give predecessor hit probabilities
$\hat{p}^-(x)$ for $x\in U^-$. These predictions also define $s^-$ and the
candidate features, with the phenotype scale and screening progress
recomputed from $D^-$. The Hellinger weight compares $\hat{p}^-(x)$ and
$\hat{p}^+(x)$ over the shared remaining pool $U^+\subseteq U^-$.
The observed discoveries in $A$ give the transition reward $R$. The
predecessor state, hidden batch, reward, weight, and successor state then
form the LBO transition.

\begin{algorithm}[!t]
\caption{LBO transition construction}
\renewcommand{\algorithmicrequire}{\textbf{Input}}
\renewcommand{\algorithmicensure}{\textbf{Output}}
\label{alg:lbo-batches}
\begin{algorithmic}[1]
\Require Measured data $D^+$, unmeasured pool $U^+$, initial data $D_0$,
successor state $s^+$, successor estimates $\{\hat{p}^+(x):x\in U^+\}$ when nonterminal,
stage, batch size $B$, threshold $\tau$, transition limit $L$,
replay buffer $\mathcal R$, and current real batch $A_t$ for online LBO.
\Ensure Replay buffer $\mathcal R$ augmented with up to $L$ LBO transitions.
\If{initial LBO}
    \State Set eligible candidates $E\leftarrow\{x:(x,e(x))\in D_0\}$.
    \State Initialize used batches $\mathcal S\leftarrow\varnothing$.
\Else
    \State Set $E\leftarrow\{x:(x,e(x))\in D^+\setminus D_0\}$.
    \State Initialize $\mathcal S\leftarrow\{A_t\}$.
\EndIf
\State Initialize the hidden batch list $\mathcal C\leftarrow[\,]$.
\State Generate seeded random permutations of $E$.
\For{each generated permutation}
    \State Partition it into complete batches of size $B$.
    \For{each batch $A$ in this partition}
        \If{$A\notin\mathcal S$}
            \State Append $A$ to $\mathcal C$.
            \State Add $A$ to $\mathcal S$.
        \EndIf
    \EndFor
\EndFor
\State Retain at most the first $L$ batches in $\mathcal C$.
\For{each hidden batch $A\in\mathcal C$}
    \State Hide its outcomes $D^-\leftarrow D^+\setminus\{(x,e(x)):x\in A\}$.
    \State Restore its candidates to the pool $U^-\leftarrow U^+\cup A$.
    \State Fit predecessor surrogate $g_{\beta^-}$ to $D^-$ with validation early stopping.
    \State Draw $K=50$ MC dropout predictions from $g_{\beta^-}$ for $x\in U^-$.
    \State Estimate $\hat{p}^-(x)\leftarrow K^{-1}\sum\nolimits_{k=1}^K\mathbf1\{g_{\beta^-}^{(k)}(\phi(x))\ge\tau\}$ for $x\in U^-$.
    \State Construct $s^-$ and $\{c^-(x):x\in U^-\}$ from $D^-$ and its predictive statistics.
    \State Compute $R\leftarrow\sum\nolimits_{x\in A}z(x)$ from the hidden outcomes.
    \If{$s^+$ is terminal}
        \State Set $w\leftarrow0$.
    \Else
        \State Compute $w$ from Hellinger distances between $\hat{p}^-(x)$ and $\hat{p}^+(x)$ for $x\in U^+$.
    \EndIf
    \State Add $\widetilde\xi=(s^-,A,R,w,s^+)$ and its prediction inputs to $\mathcal R$.
\EndFor
\State \Return $\mathcal R$.
\end{algorithmic}
\end{algorithm}

The set $\mathcal S$ prevents repeated hidden batches for a given successor
and excludes the current real batch during online LBO. The list
$\mathcal C$ contains the accepted hidden batches. Initial and online
transition limits $L$ and the generation schedule are specified in
Table~\ref{tab:tdal-hyperparameters}.

\textbf{Successor selection and TD target.}
For a replay transition $\xi_i=(s_i,A_i,R_i,w_i,s'_i)$, the TD target
combines its observed reward with an estimate of weighted future discoveries.
To estimate this successor value, the current acquisition function scores
every candidate in the recorded successor pool $U'_i$ using
\begin{equation}
q_\theta(s'_i,x)=\hat{p}'_i(x)+f_\theta(s'_i,c'_i(x)),
\qquad x\in U'_i.
\end{equation}
The state $s'_i$, candidate features $c'_i(x)$, and hit probabilities
$\hat{p}'_i(x)$ are saved when the transition is constructed. Replay keeps
these inputs fixed and applies the current network to them. As $\theta$
changes, the network can select a different successor batch from the same
recorded pool.

Following Double DQN, we select the $B$ candidates with the highest current
scores to form $A'_{\theta,i}$ and evaluate them with the target network
\citep{c10_vanhasselt2016doubleq}. This gives
\begin{equation}
\begin{aligned}
Q_{\bar\theta}(s'_i,A'_{\theta,i})
&=\sum\nolimits_{x\in A'_{\theta,i}}
\left[\hat{p}'_i(x)+f_{\bar\theta}(s'_i,c'_i(x))\right],\\
Y_i&=R_i+w_iQ_{\bar\theta}(s'_i,A'_{\theta,i}).
\end{aligned}
\end{equation}
This successor selection is used to calculate the training target and does
not trigger additional experiments. For terminal transitions, $Y_i=R_i$.
Real and LBO transitions use the same target, which is held constant when
computing gradients.

\textbf{Experience replay and parameter updates.}
Real and LBO transitions enter a shared prioritized replay buffer.
Let $n$ be the number of stored transitions and $\rho_i$ the priority of
transition $i$. Each update samples a training batch $\mathcal B$ without
replacement using normalized sampling weights
\begin{equation}
P(i)=\frac{(\rho_i+\epsilon_{\mathrm{PER}})^{\alpha_{\mathrm{PER}}}}
{\sum\nolimits_{j=1}^{n}(\rho_j+\epsilon_{\mathrm{PER}})^{\alpha_{\mathrm{PER}}}}.
\end{equation}
The exponent $\alpha_{\mathrm{PER}}$ controls the preference for high
priority transitions. Frequently sampled transitions receive smaller importance weights in the
loss, reducing the imbalance introduced by prioritized sampling. We use
\begin{equation}
\widetilde\kappa_i=(nP(i))^{-\beta_{\mathrm{IS}}},
\qquad
\kappa_i^{\mathrm{IS}}=
\frac{\widetilde\kappa_i}{\max\nolimits_{j\in\mathcal B}\widetilde\kappa_j}.
\end{equation}
Here, $\beta_{\mathrm{IS}}$ controls the strength of the weighting, and
normalization sets the largest weight within $\mathcal B$ to one.
With TD error $\delta_i=Q_\theta(s_i,A_i)-Y_i$, the training loss is
\begin{equation}
\widehat{\mathcal L}_{\mathrm{TD}}(\theta)
=\frac{1}{|\mathcal B|}\sum\nolimits_{i\in\mathcal B}
\kappa_i^{\mathrm{IS}}\operatorname{Huber}_1(\delta_i).
\end{equation}
The Huber loss is quadratic for absolute errors up to 1 and linear above 1.
Gradients update only $\theta$, with surrogate predictions held fixed.
After each update, the sampled transitions receive priorities
$\rho_i\leftarrow|\delta_i|+\epsilon_{\mathrm{PER}}$, and the target parameters
follow
\begin{equation}
\bar\theta\leftarrow(1-\tau_{\mathrm{target}})\bar\theta
+\tau_{\mathrm{target}}\theta.
\end{equation}
The optimizer and training settings are specified in
Table~\ref{tab:tdal-hyperparameters}.

\section{Theoretical Analysis}
\label{app:theory}

We prove Propositions 1--3, establishing weighted Bellman optimality,
relating the Hellinger weight to predictive improvement, and bounding
the gap in final discoveries.

\subsection{Weighted Bellman Optimality}
\label{app:weighted-bellman-proof}

We fix the surrogate training and prediction procedures and condition on their
randomness. Given the history $H_t$ and query batch $A_t$, the observed outcomes
determine the discovery count $R_t$ and feedback weight
$w_t=w(H_t,A_t,H_{t+1})\in[0,1]$. Both quantities are therefore determined by
the successor history $H_{t+1}$. At history $H$, the feasible batch set is
\begin{equation}
\mathcal A_t(H)=\{A\subseteq U_t(H):|A|=B\}.
\end{equation}

For any selection policy $\pi$, define the weighted discovery count
from round $t$ to the end of screening as
\begin{equation}
G_t^w=\sum\nolimits_{k=t}^{T-1}
\left(\prod\nolimits_{j=t}^{k-1}w_j\right)R_k,
\qquad G_T^w=0.
\end{equation}
The current count $R_t$ has weight one, while each subsequent count is
weighted by the product of the intervening feedback weights.
The corresponding policy value is
\begin{equation}
V_t^{w,\pi}(H)=\mathbb E_\pi[G_t^w\mid H_t=H],
\end{equation}
the expected weighted number of remaining discoveries when selection
follows $\pi$ from the current screening history $H$.
The terminal value is zero because no experiments remain.

Separating the current round from subsequent rounds gives
\begin{equation}
G_t^w=R_t+w_tG_{t+1}^w.
\end{equation}
Given $H_{t+1}$, both $R_t$ and $w_t$ are known, and the conditional
expectation of the remaining return is $V_{t+1}^{w,\pi}(H_{t+1})$.
Averaging over the current batch and its observations, we obtain the following by the law of iterated expectations:
\begin{equation}
\label{eq:weighted-policy-evaluation}
V_t^{w,\pi}(H)
=\sum\nolimits_{A\in\mathcal A_t(H)}\pi_t(A\mid H)
\mathbb E\!\left[
R_t+w_tV_{t+1}^{w,\pi}(H_{t+1})
\mid H_t=H,A_t=A
\right].
\end{equation}
Here, $\pi_t(A\mid H)$ is the probability of selecting batch $A$ given
history $H$. This expression averages the expected current discoveries
and weighted successor value over the policy's batch choices.

Starting from the terminal condition $V_T^{*,w}=0$, define recursively
\begin{equation}
\label{eq:weighted-optimal-recursion}
\begin{aligned}
Q_t^{*,w}(H,A)
&=\mathbb E\!\left[
R_t+w_tV_{t+1}^{*,w}(H_{t+1})
\mid H_t=H,A_t=A
\right],\\
V_t^{*,w}(H)&=\max\nolimits_{A\in\mathcal A_t(H)}Q_t^{*,w}(H,A).
\end{aligned}
\end{equation}
This recursion starts at the last round, where only current discoveries
contribute. Earlier rounds also include the weighted successor value
already determined at the next round.
Here, $Q_t^{*,w}(H,A)$ is the conditional expected value of selecting
batch $A$, and $V_t^{*,w}(H)$ takes the maximum over feasible batches.
The finite horizon and finite batch sets uniquely determine these values.
Since $0\le R_t\le B$ and $0\le w_t\le1$,
\begin{equation}
0\le Q_t^{*,w}(H,A)\le V_t^{*,w}(H)\le(T-t)B.
\end{equation}

To establish optimality, we first show that every policy has value at
most $V_t^{*,w}$. At the terminal round $t=T$, every policy satisfies
\begin{equation}
V_T^{w,\pi}(H)=V_T^{*,w}(H)=0.
\end{equation}
Suppose that, at round $t+1$, every policy and reachable history satisfy
\begin{equation}
V_{t+1}^{w,\pi}(H_{t+1})\le V_{t+1}^{*,w}(H_{t+1}).
\end{equation}
At round $t$, retain the policy's current batch selection probabilities
and replace its successor value by this upper bound.
Since $w_t\ge0$, this replacement cannot decrease the conditional
expected value. Equations~\ref{eq:weighted-policy-evaluation}
and~\ref{eq:weighted-optimal-recursion} therefore give
\begin{equation}
\begin{aligned}
V_t^{w,\pi}(H)
&\le\sum\nolimits_{A\in\mathcal A_t(H)}\pi_t(A\mid H)
\mathbb E\!\left[
R_t+w_tV_{t+1}^{*,w}(H_{t+1})
\mid H_t=H,A_t=A
\right]\\
&=\sum\nolimits_{A\in\mathcal A_t(H)}\pi_t(A\mid H)Q_t^{*,w}(H,A)\\
&\le V_t^{*,w}(H).
\end{aligned}
\end{equation}
The last inequality holds because any probability-weighted average of
batch values cannot exceed their maximum.
Thus, the value bound at the next round implies the bound at the current
round. Backward induction from the terminal round establishes
$V_t^{w,\pi}(H)\le V_t^{*,w}(H)$ for every round, policy, and reachable
history.

We next show that this upper bound is attained. Let $\pi_w^*$ select a
feasible batch maximizing $Q_t^{*,w}$ at every round and reachable history.
At the terminal round,
\begin{equation}
V_T^{w,\pi_w^*}(H)=V_T^{*,w}(H)=0.
\end{equation}
Suppose this policy attains $V_{t+1}^{*,w}$ at every reachable history at
round $t+1$. After selecting $\pi_w^*(H)$ at round $t$, its continuation
therefore attains the optimal successor value in the definition of $Q_t^{*,w}$.
\begin{equation}
\begin{aligned}
V_t^{w,\pi_w^*}(H)
&=Q_t^{*,w}\bigl(H,\pi_w^*(H)\bigr)\\
&=\max\nolimits_{A\in\mathcal A_t(H)}Q_t^{*,w}(H,A)\\
&=V_t^{*,w}(H).
\end{aligned}
\end{equation}
The first equality follows from the induction hypothesis, the second from
the policy's maximizing batch choice, and the third from the definition of
$V_t^{*,w}$. Backward induction therefore establishes that this policy
attains the bound at every round and reachable history.

Finally, since $J_w(\pi)=V_0^{w,\pi}(H_0)$, every policy's objective is bounded
above by $V_0^{*,w}(H_0)$, and $\pi_w^*$ attains this value. Therefore,
\begin{equation}
\begin{aligned}
J_w(\pi_w^*)
&=V_0^{w,\pi_w^*}(H_0)=V_0^{*,w}(H_0)\\
&=\max\nolimits_{A\in\mathcal A_0(H_0)}Q_0^{*,w}(H_0,A)
=\max\nolimits_\pi J_w(\pi).
\end{aligned}
\end{equation}
Thus, maximizing the optimal batch value at each round maximizes the
weighted cumulative discovery objective. This proves Proposition 1.

\subsection{Feedback and Predictive Improvement}
\label{app:predictive-learning}

Fix the current history $H_t$ and query batch $A_t$, with remaining
pool $U=U_{t+1}$ and $|U|>0$. Under a common joint probability model,
define the conditional hit probabilities before and after feedback as
\begin{equation}
\begin{aligned}
p_x&=\Pr(z(x)=1\mid H_t,A_t),\\
q_x&=\Pr(z(x)=1\mid H_t,A_t,O_t).
\end{aligned}
\end{equation} We analyze the Hellinger weight computed
from these probabilities.
Before the batch outcomes are observed, averaging the updated probability
$q_x$ over their possible values recovers the current probability $p_x$, giving
\begin{equation}
\mathbb E[q_x\mid H_t,A_t]=p_x.
\end{equation}
This analysis uses conditional hit probabilities before and after experimental
feedback. TDAL approximates these probabilities using MC dropout predictions.

The expected reduction in mean squared prediction error is
\begin{equation}
\Delta_{\mathrm{pred}}
=\frac1{|U|}\sum\nolimits_{x\in U}
\mathbb E\!\left[(p_x-z(x))^2-(q_x-z(x))^2\right].
\end{equation}
The two terms are the binary Brier losses before and after feedback
\citep{c15_gneiting2007proper}. Expanding their difference gives
\begin{equation}
(p_x-z(x))^2-(q_x-z(x))^2
=(q_x-p_x)^2+2(q_x-p_x)(z(x)-q_x).
\end{equation}
By the definition of $q_x$,
\begin{equation}
\mathbb E[z(x)-q_x\mid H_t,A_t,O_t]
=\Pr(z(x)=1\mid H_t,A_t,O_t)-q_x=0.
\end{equation}
Given the feedback, $q_x-p_x$ is fixed, so the cross term has zero
conditional expectation. The law of iterated expectations therefore yields
\begin{equation}
\label{eq:predictive-learning-identity}
\Delta_{\mathrm{pred}}
=\frac1{|U|}\sum\nolimits_{x\in U}\mathbb E[(q_x-p_x)^2].
\end{equation}
Thus, feedback does not increase mean prediction error in expectation,
and the expected reduction equals the mean squared change in hit probabilities.

To relate this reduction to the Hellinger weight, write
\begin{equation}
\begin{aligned}
h_x^2&=1-\sqrt{p_xq_x}-\sqrt{(1-p_x)(1-q_x)},\\
w_t^2&=\frac1{|U|}\sum\nolimits_{x\in U}h_x^2.
\end{aligned}
\end{equation}
For any $p,q\in[0,1]$, let $h=H_{\mathrm{Ber}}(p,q)$.
The Bernoulli Hellinger bounds \citep{c16_sason2016divergence}
follow directly here. Since $\sqrt{ab}\ge\min(a,b)$,
\begin{equation}
h^2\le1-\min(p,q)-\min(1-p,1-q)=|p-q|.
\end{equation}
Conversely, define
\begin{equation}
u=(\sqrt p,\sqrt{1-p}),\qquad v=(\sqrt q,\sqrt{1-q}).
\end{equation}
Cauchy--Schwarz gives
\begin{equation}
\begin{aligned}
|p-q|
&=\frac12\sum\nolimits_{i=1}^2|u_i-v_i|\,|u_i+v_i|\\
&\le\frac12\|u-v\|_2\|u+v\|_2
=h\sqrt{2-h^2}\le\sqrt2\,h.
\end{aligned}
\end{equation}
Hence
\begin{equation}
h_x^4\le(q_x-p_x)^2\le2h_x^2.
\end{equation}
Jensen's inequality gives, for each feedback outcome,
\begin{equation}
\label{eq:predictive-learning-distance}
\begin{aligned}
w_t^4
&=\left(\frac1{|U|}\sum\nolimits_{x\in U}h_x^2\right)^2\\
&\le\frac1{|U|}\sum\nolimits_{x\in U}h_x^4\\
&\le\frac1{|U|}\sum\nolimits_{x\in U}(q_x-p_x)^2
\le2w_t^2.
\end{aligned}
\end{equation}
Taking conditional expectations gives
\begin{equation}
\mathbb E[w_t^4]\le\Delta_{\mathrm{pred}}\le2\mathbb E[w_t^2].
\end{equation}
An experiment therefore does more than establish whether queried candidates
are hits. It can improve predictions for unmeasured candidates. These bounds
connect that expected improvement to the Hellinger weight, supporting the use
of feedback's effect on remaining predictions to weight successor value.

Now consider querying one candidate $a$ and observing only its hit indicator.
Suppress conditioning on the current history and query, and define
\begin{equation}
\begin{aligned}
r&=\Pr(z(a)=1)\in(0,1),\\
p_x^+&=\Pr(z(x)=1\mid z(a)=1),\\
p_x^-&=\Pr(z(x)=1\mid z(a)=0).
\end{aligned}
\end{equation}
The law of total probability gives
\begin{equation}
p_x=rp_x^++(1-r)p_x^-.
\end{equation}
After feedback, $q_x=p_x^+$ with probability $r$ and $q_x=p_x^-$
with probability $1-r$. Hence
\begin{equation}
\begin{aligned}
\mathbb E[(q_x-p_x)^2]
&=r(p_x^+-p_x)^2+(1-r)(p_x^--p_x)^2\\
&=r(1-r)^2(p_x^+-p_x^-)^2+(1-r)r^2(p_x^+-p_x^-)^2\\
&=r(1-r)(p_x^+-p_x^-)^2.
\end{aligned}
\end{equation}
Averaging over the remaining candidates yields
\begin{equation}
\label{eq:binary-feedback-learning}
\Delta_{\mathrm{pred}}(a)
=\frac{r(1-r)}{|U|}\sum\nolimits_{x\in U}(p_x^+-p_x^-)^2.
\end{equation}
This proves Proposition 2.

Outcome uncertainty and its influence on remaining candidates jointly
determine predictive improvement. If neither outcome changes remaining
predictions, the feedback provides no predictive improvement. Otherwise,
even a query without a hit changes remaining predictions and produces a
nonzero weight. This explains how TDAL retains the current discovery reward
while allowing experiments without hits to contribute to learning through
successor value.

The decomposition into two outcomes is the special case of one query revealing
only a binary label. TDAL observes a batch of continuous phenotypes instead.
The general identity for predictive improvement conditions on the complete
batch observations $O_t$ and also applies to that feedback under the same
assumptions about the joint model. It does not require a binary observation space.

\subsection{Final Discovery Gap}
\label{app:final-discovery-gap}

Fix the initial history $H_0$, surrogate update rule, and algorithmic randomness.
At each round, $0\le R_t\le B$, and the feedback weight is a measurable
function $w_t=w(H_t,A_t,H_{t+1})\in[0,1]$. This permits the weight,
current discovery count, and successor value to depend on the same feedback.
Write
\begin{equation}
W_t=\prod\nolimits_{j=0}^{t-1}w_j,\qquad W_0=1.
\end{equation}

At each history $H$, let $s=s_t(H)$ be the screening state and let
$\theta$ denote the network parameters after all online updates at that history.
The learned policy selects a feasible batch maximizing $Q_\theta(s,A)$.
Its value estimation error is
\begin{equation}
\varepsilon_t^w(H)
=\max\nolimits_{A\in\mathcal A_t(H)}
|Q_\theta(s,A)-Q_t^{*,w}(H,A)|.
\end{equation}
First consider how this error affects the current batch choice.
Fix a history $H$, let $A^*$ maximize $Q_t^{*,w}(H,A)$,
and let $\widehat A$ maximize $Q_\theta(s,A)$.
By the error definition,
\begin{equation}
\begin{aligned}
V_t^{*,w}(H)
&=Q_t^{*,w}(H,A^*)\\
&\le Q_\theta(s,A^*)+\varepsilon_t^w(H)\\
&\le Q_\theta(s,\widehat A)+\varepsilon_t^w(H)\\
&\le Q_t^{*,w}(H,\widehat A)+2\varepsilon_t^w(H).
\end{aligned}
\end{equation}
The second inequality uses the maximizing property of $\widehat A$.
Thus,
\begin{equation}
\label{eq:batch-value-gap}
0\le V_t^{*,w}(H)-Q_t^{*,w}(H,\widehat A)
\le2\varepsilon_t^w(H).
\end{equation}
The gap between the optimal batch value and the value of the selected
batch is therefore at most twice the estimation error.

Next, accumulate these losses across the screening trajectory.
Let $\widehat A_t$ be the batch selected by the learned policy at round $t$.
The weighted Bellman equation gives
\begin{equation}
Q_t^{*,w}(H_t,\widehat A_t)
=\mathbb E\!\left[
R_t+w_tV_{t+1}^{*,w}(H_{t+1})
\mid H_t,A_t=\widehat A_t\right].
\end{equation}
Since $W_t$ is determined by the current history and $W_{t+1}=W_tw_t$,
taking expectations along the learned policy gives
\begin{equation}
\label{eq:weighted-performance-difference}
\begin{aligned}
&\mathbb E_{\pi_\theta}\!\left[
\sum\nolimits_{t=0}^{T-1}W_t
\bigl(V_t^{*,w}(H_t)-Q_t^{*,w}(H_t,\widehat A_t)\bigr)
\mid H_0\right]\\
&\quad=\mathbb E_{\pi_\theta}\!\left[
\sum\nolimits_{t=0}^{T-1}
\bigl(W_tV_t^{*,w}(H_t)-W_tR_t-W_{t+1}V_{t+1}^{*,w}(H_{t+1})\bigr)
\mid H_0\right]\\
&\quad=V_0^{*,w}(H_0)-J_w(\pi_\theta).
\end{aligned}
\end{equation}
Value terms from adjacent rounds cancel, leaving the initial value
$V_0^{*,w}(H_0)$. The terminal term vanishes because $V_T^{*,w}=0$.
Proposition 1 and the batch value gap then imply
\begin{equation}
\label{eq:weighted-learning-gap}
\max\nolimits_\pi J_w(\pi)-J_w(\pi_\theta)
\le2\mathbb E_{\pi_\theta}\!\left[
\sum\nolimits_{t=0}^{T-1}W_t\varepsilon_t^w(H_t)\mid H_0\right].
\end{equation}

Finally, relate the weighted objective to final discoveries.
Since
\begin{equation}
C_T=C_0+\sum\nolimits_{t=0}^{T-1}R_t,
\end{equation}
every policy satisfies
\begin{equation}
\mathbb E_\pi[C_T\mid H_0]
=C_0+J_w(\pi)
+\mathbb E_\pi\!\left[
\sum\nolimits_{t=0}^{T-1}(1-W_t)R_t\mid H_0\right].
\end{equation}
The last term counts the expected discoveries omitted when each round
receives weight $W_t$ instead of one. It is nonnegative because $R_t\ge0$
and $W_t\le1$.
Let $\pi_C^*$ maximize expected final discoveries under the same initial
information and budget, and write
\begin{equation}
\Delta_C(w)=\mathbb E_{\pi_C^*}\!\left[
\sum\nolimits_{t=0}^{T-1}(1-W_t)R_t\mid H_0\right].
\end{equation}
Subtracting the final discovery counts and using the nonnegative
weighting bias of the learned policy gives
\begin{equation}
\begin{aligned}
&\mathbb E_{\pi_C^*}[C_T\mid H_0]
-\mathbb E_{\pi_\theta}[C_T\mid H_0]\\
&\quad\le\Delta_C(w)+J_w(\pi_C^*)-J_w(\pi_\theta)\\
&\quad\le\Delta_C(w)+\max\nolimits_\pi J_w(\pi)-J_w(\pi_\theta)\\
&\quad\le\Delta_C(w)+2\mathbb E_{\pi_\theta}\!\left[
\sum\nolimits_{t=0}^{T-1}W_t\varepsilon_t^w(H_t)\mid H_0\right].
\end{aligned}
\end{equation}
This proves Proposition 3.

\textbf{Feedback weights and the two error terms.}
Consider two weight rules $w,w'$ in the same experimental environment.
If, at every reachable history, batch, and feedback outcome,
\begin{equation}
w'(H,A,H')\ge w(H,A,H'),
\end{equation}
then $W_t'\ge W_t$ on every trajectory.
Taking expectations under the same optimal discovery policy $\pi_C^*$ yields
\begin{equation}
\Delta_C(w')\le\Delta_C(w).
\end{equation}
Thus, increasing feedback weights reduces the objective weighting bias.

To examine the error term, consider constant weights $w_t=w$ and assume
all weight settings being compared share a common value error bound
\begin{equation}
\varepsilon_t^w(H)\le\varepsilon.
\end{equation}
Then $W_t=w^t$, and $R_t\le B$ gives
\begin{equation}
\Delta_C(w)\le B\sum\nolimits_{t=0}^{T-1}(1-w^t),
\end{equation}
while
\begin{equation}
2\mathbb E_{\pi_\theta}\!\left[
\sum\nolimits_{t=0}^{T-1}W_t\varepsilon_t^w(H_t)\mid H_0\right]
\le2\varepsilon\sum\nolimits_{t=0}^{T-1}w^t.
\end{equation}
The bias bound is nonincreasing in $w$, whereas the estimation error
term's bound is nondecreasing under the common error bound $\varepsilon$.
Smaller weights reduce the error bound but attenuate more future discovery
return. Larger weights retain more future return but increase the error bound.
In the learned model, $\varepsilon_t^w(H)$
and the selected trajectory depend on $w$. Comparing weight settings
therefore also involves their different value estimation errors.
When $w=1$, the objective weighting bias vanishes and the final discovery
gap is at most $2T\varepsilon$. If also $\varepsilon=0$, the learned policy
achieves the optimal expected final discovery count.

\section{Datasets and Experimental Setup}
\label{app:experimental-settings}
\label{app:comparison-protocol}

\renewcommand{\thetable}{\thesection\arabic{table}}
\renewcommand{\theHtable}{\thesection.\arabic{table}}
\setcounter{table}{0}

\subsection{Datasets and Hit Definitions}
\label{app:datasets-hits}
The eight public screens comprise four single gene tasks, three double gene
tasks, and one triple gene task, covering T cell function, cancer
cell growth, immune killing, and genetic interactions. Each candidate is
a single gene or combinatorial perturbation. Table~\ref{tab:phenotypes-hits}
summarizes perturbation types, phenotypes, and hit rules.

\begin{table}[htbp]
\centering
\caption{Perturbation types, phenotypes, and hit rules for eight tasks.}
\label{tab:phenotypes-hits}
\small
\setlength{\tabcolsep}{4pt}
\renewcommand{\arraystretch}{1.15}
\begin{tabular}{@{}>{\raggedright\arraybackslash}p{0.25\linewidth}>{\raggedright\arraybackslash}p{0.12\linewidth}>{\raggedright\arraybackslash}p{0.38\linewidth}>{\raggedright\arraybackslash}p{0.18\linewidth}@{}}
\toprule
Dataset & Perturbation & Phenotype & Hit rule \\
\midrule
IFN-$\gamma$ \citep{b01_schmidt2022crispr} & Single gene & IFN-$\gamma$ production by T cells & Top 5\% by absolute score \\
\addlinespace[1.5pt]
IL-2 \citep{b01_schmidt2022crispr} & Single gene & IL-2 production by T cells & Top 5\% by absolute score \\
\addlinespace[1.5pt]
NK \citep{b02_zhuang2019genome} & Single gene & Cancer cell sensitivity to NK cell killing & Top 5\% by absolute score \\
\addlinespace[1.5pt]
Carnevale \citep{b16_carnevale2022rasa2} & Single gene & T cell proliferation under adenosine suppression & Top 5\% by absolute score \\
\addlinespace[1.5pt]
Simpson K562 \citep{b25_simpson2023parp} & Two genes & K562 cell growth after dual gene repression & Top 5\% by raw score \\
\addlinespace[1.5pt]
SCHEMATIC \citep{b26_fong2025multilineage} & Two genes & Genetic interactions across cancer cell backgrounds & Bottom 5\% by raw score \\
\addlinespace[1.5pt]
Kuzmin yeast \citep{b24_kuzmin2020duplicate} & Three genes & Genetic interactions in triple gene yeast perturbations & Bottom 5\% by raw score \\
\addlinespace[1.5pt]
Horlbeck Jurkat \citep{b17_horlbeck2018mapping} & Two genes & Jurkat cell growth after dual gene repression & Bottom 5\% by raw score \\
\addlinespace[1.5pt]
\bottomrule
\end{tabular}
\end{table}

Raw phenotype scores are transformed into effects $e(x)$ according to the
direction in the table. For a pool of $N=|\mathcal X|$ candidates, the score
at rank $\max\{1,\lfloor0.05N\rfloor\}$ in descending effect order defines
the fixed threshold $\tau$, with hits given by
$z(x)=\mathbf 1\{e(x)\ge\tau\}$. The target hit proportion is 5\%, and
$\tau$ is the corresponding phenotype score threshold. The threshold is
fixed before screening. The phenotype and hit label of each candidate are
revealed when it is queried.

\textbf{IFN-$\gamma$.}
This task identifies genes that regulate interferon-$\gamma$ production in
T cells. Enrichment differences between populations with high and low cytokine levels
measure perturbation effects. We target strong effects in either direction
\citep{b01_schmidt2022crispr}.

\textbf{IL-2.}
This task identifies regulators of interleukin-2 production following T cell
stimulation. Enrichment differences between populations with high and low expression define the phenotype, with strong effects in either direction
included as discovery targets \citep{b01_schmidt2022crispr}.

\textbf{NK.}
This task studies cancer cell genes that affect sensitivity or resistance
to natural killer cells. The phenotype is the sum of standardized enrichment
scores across two replicates for perturbed K562 cells exposed to NK cells
relative to controls. Strong effects in either direction are targeted
\citep{b02_zhuang2019genome}.

\textbf{Carnevale.}
This task studies genes that affect T cell proliferation under
immunosuppressive conditions. We use enrichment scores associated with proliferation
from the screen under adenosine suppression and target perturbations that
strongly increase or decrease proliferation \citep{b16_carnevale2022rasa2}.

\textbf{Simpson K562.}
The original study examines how DNA repair genes jointly affect cell growth
and PARP inhibitor response. We use its double gene CRISPRi growth phenotype
and target combinations with the highest growth scores
\citep{b25_simpson2023parp}.

\textbf{SCHEMATIC.}
This task seeks gene combinations that cause synthetic lethality across cancer cell
backgrounds. We use the multilineage interaction score to identify strong
negative interactions whose growth or survival defects exceed expectations
from individual perturbations \citep{b26_fong2025multilineage}.

\textbf{Kuzmin yeast.}
This task examines how duplicate genes compensate for one another and how
this compensation depends on other genes. We use the corrected trigenic
interaction score to target combinations with stronger growth defects than
expected \citep{b24_kuzmin2020duplicate}.

\textbf{Horlbeck Jurkat.}
This task examines the effect of jointly repressing two genes on Jurkat
cell growth. We use the double gene CRISPRi growth phenotype and target
combinations that strongly inhibit growth \citep{b17_horlbeck2018mapping}.

\subsection{Experimental Protocol and Hyperparameters}
\label{app:experimental-protocol}
Table~\ref{tab:data-budgets} summarizes candidate counts, pool hit counts,
and screening configurations.

\begin{table}[htbp]
    \centering
    \caption{Candidate pools, hits, and budgets for eight tasks.}
    \label{tab:data-budgets}
    \setlength{\tabcolsep}{4pt}
    \renewcommand{\arraystretch}{1.18}
    \begin{tabular*}{\linewidth}{@{\extracolsep{\fill}}lrrrrrr@{}}
        \toprule
        Dataset & $N$ & Pool hits & $n_0$ & $B$ & $T$ & $M$ \\
        \midrule
        IFN-$\gamma$ & 17,470 & 873 & 128 & 32 & 60 & 2,048 \\
        IL-2 & 17,470 & 873 & 128 & 32 & 60 & 2,048 \\
        NK & 17,533 & 876 & 128 & 32 & 60 & 2,048 \\
        Carnevale & 17,526 & 876 & 128 & 32 & 60 & 2,048 \\
        Simpson K562 & 143,345 & 7,167 & 1,024 & 128 & 200 & 26,624 \\
        SCHEMATIC & 11,067 & 553 & 128 & 32 & 60 & 2,048 \\
        Kuzmin yeast & 214,642 & 10,732 & 1,024 & 256 & 200 & 52,224 \\
        Horlbeck Jurkat & 91,378 & 4,568 & 1,024 & 128 & 200 & 26,624 \\
        \bottomrule
    \end{tabular*}
\end{table}

$N$ is the number of queryable perturbations with valid measured phenotypes
and fixed representations in the retrospective evaluation. Combinatorial
tasks count gene combinations as candidates. All candidate identities and
representations are known at the start. Each method initially receives
the phenotypes of the $n_0$ candidates in $D_0$, while the phenotypes of
the remaining $N-n_0$ candidates are withheld. After batch $A_t$ is selected,
the evaluation environment returns its recorded experimental outcomes,
adds these observations to the measured data, and removes the candidates
from the unmeasured pool. Screening returns all verified hit candidates,
whose count $C_T$ measures final discovery.

For seed $i$, let $C_{t,i}$ be the number of hits in the measured set after
$t$ active rounds, with $C_{0,i}$ the number in the initial set. The final
cumulative hit count is
\begin{equation}
\label{eq:appendix-final-hits}
C_{T,i}=C_{0,i}+\sum_{t=0}^{T-1}R_{t,i}.
\end{equation}
This metric counts distinct discovered hits within the total budget,
including initialization. Recall is $C_{T,i}$ divided by the fixed pool hit
count in Table~\ref{tab:data-budgets}. Batch size $B$ gives the number of new
queries per round, $T$ is the number of active rounds, and the total budget
$M=n_0+BT$ includes initialization and all subsequent queries.

Each task is repeated with seeds 0--4. Methods share the candidate pool,
initial set, hit threshold, experimental budget, and underlying phenotype dataset within each seed. Method differences are paired within each task and seed
before aggregation across seeds.

\Needspace{4\baselineskip}
\textbf{Hyperparameter settings.}
Table~\ref{tab:tdal-hyperparameters} summarizes TDAL's model and training settings.

\begin{table}[!htbp]
\centering
\small
\setlength{\tabcolsep}{4pt}
\renewcommand{\arraystretch}{1.18}
\caption{Key TDAL hyperparameters.}
\label{tab:tdal-hyperparameters}
\begin{tabular}{@{}>{\raggedright\arraybackslash}p{0.17\linewidth}>{\raggedright\arraybackslash}p{0.29\linewidth}>{\raggedright\arraybackslash}p{0.49\linewidth}@{}}
\toprule
Component & Parameter & Value \\
\midrule
Surrogate & Hidden widths and dropout & Widths 128, 64, dropout probability 0.2 \\
\addlinespace[1.5pt]
Surrogate & Training/validation split & 80\%/20\% \\
\addlinespace[1.5pt]
Surrogate & Optimizer & Adam, learning rate $10^{-3}$ \\
\addlinespace[1.5pt]
Surrogate & Training batch and epochs & Batch size 64, at most 100 epochs, early stopping patience 10 \\
\addlinespace[1.5pt]
Surrogate & MC samples & $K=50$ \\
\addlinespace[1.5pt]
Acquisition network & Projection and hidden widths & Projection dimension 64, hidden widths 128, 128 \\
\addlinespace[1.5pt]
TD learning & Optimizer & AdamW, learning rate $5\times10^{-4}$ \\
\addlinespace[1.5pt]
TD learning & Training batch and updates & At most 16 transitions per update, 20 updates initially and after each active round \\
\addlinespace[1.5pt]
Initial LBO & Transition limit & $\min\{80,\,2\lfloor\lvert D_0\rvert/B\rfloor\}$ \\
\addlinespace[1.5pt]
Online LBO & Rounds 1--8/9--32/33--64 & At most 8/4/2 transitions per round \\
\addlinespace[1.5pt]
Online LBO & From round 65 & At most 1 transition every 4 rounds \\
\addlinespace[1.5pt]
\bottomrule
\end{tabular}
\end{table}

\subsection{Baseline Methods}
\label{app:baseline-methods}
For batch perturbation screening, we compare representative methods for
hit discovery, Bayesian optimization, and batch active learning, with
random selection as a reference.
UCB, EI, TS, TopUncertain, and OAE use the GeneDisco Bayesian MLP
and retrain their surrogate after each round of new observations
\citep{p02_mehrjou2022genedisco}.

\textbf{PoH.}
Many Needles selects the $B$ unmeasured candidates with the highest estimated
probability of exceeding the hit threshold \citep{p01_rubbi2026many}.
At each round, we refit a Bayesian MLP with two hidden layers of widths 128
and 64 on the measured data. For each candidate $x\in U_t$, we obtain 50 MC
dropout predictions and use their mean and standard deviation to form a
Gaussian predictive approximation. The probability of exceeding $\tau$
then serves as the acquisition score.

\textbf{OAE.}
OAE uses optimistic estimates for batch discovery of candidates with high responses. We use its released Bayesian variant, which combines predictive
means and uncertainty to select each batch after aligning predictions with
the effect direction \citep{p07_pacchiano2023neural}.

\textbf{DiscoBAX.}
DiscoBAX targets a diverse subset of candidates with high responses and uses
fantasized observations to estimate information gain about that subset.
We use the authors' surrogate training and acquisition implementation,
with two hidden layers of 128 and 64 units and dropout 0.2 to match
TDAL and PoH in network capacity. We evaluate this implementation on
the four single gene perturbation tasks
\citep{p08_lyle2023discobax}.

\textbf{UCB.}
Upper Confidence Bound combines the predictive mean with a weighted
uncertainty term. It favors candidates with high predicted effects or high
uncertainty to balance exploitation and exploration
\citep{c04_srinivas2010gaussian}.

\textbf{Expected Improvement (EI).}
EI estimates the expected improvement over the best effect observed so far
and prioritizes candidates with larger expected gains
\citep{c02_jones1998efficient}.

\textbf{Thompson Sampling (TS).}
TS draws an effect function from the surrogate posterior and selects a
batch by ranking unmeasured candidates under that draw, allowing model
uncertainty to influence selection \citep{c14_thompson1933likelihood}.

\textbf{TopUncertain.}
TopUncertain ranks candidates by predictive standard deviation and allocates
experiments to regions where the surrogate is most uncertain
\citep{p02_mehrjou2022genedisco}.

\textbf{BADGE.}
BADGE selects diverse batches using k-means++ in a gradient representation
space. We use the GeneDisco regression implementation and its proxy gradient
representations \citep{c08_ash2020badge,p02_mehrjou2022genedisco}.

\textbf{CoreSet.}
CoreSet greedily expands geometric coverage in the candidate representation
space extracted by the surrogate. It successively selects the candidate
farthest from the existing selection centers
\citep{c06_sener2018coreset}.

\textbf{KMeans-Embedding.}
This method uses k-means to cluster candidate representations extracted by
the surrogate and selects candidates near cluster centers to represent different
regions of the candidate space \citep{p02_mehrjou2022genedisco}.

\textbf{Random.}
Random samples uniformly without replacement from the current unmeasured
pool, providing a reference for random screening.

\Needspace{7\baselineskip}
\section{Complete Benchmark Results}
\label{app:detailed-comparison}
\renewcommand{\thetable}{\thesection\arabic{table}}
\renewcommand{\theHtable}{\thesection.\arabic{table}}
\renewcommand{\thefigure}{\thesection\arabic{figure}}
\renewcommand{\theHfigure}{\thesection.\arabic{figure}}
\setcounter{table}{0}
\setcounter{figure}{0}

% Center two-line results naturally, with separate intra-cell and row spacing.
\newcommand{\benchmarkcell}[2]{%
    \begin{tabular}[c]{@{}c@{}}#1\\[2pt]#2\end{tabular}%
}

\subsection{Single Gene Perturbation Results}

Table~\ref{tab:single-task-final-hits} reports the complete endpoint results
for the four single gene perturbation tasks in
Figure~\ref{fig:discovery-performance}a--d, supplementing the discovery
trajectories in the main text with cumulative hit counts for each method,
variation across seeds, and recall.

\begingroup
\small
\setlength{\tabcolsep}{3pt}
    \renewcommand{\arraystretch}{1.10}
\setlength{\LTleft}{\fill}
\setlength{\LTright}{\fill}
\setlength{\LTcapwidth}{\linewidth}
\begin{longtable}{@{}lcccc@{}}
\caption{Final cumulative hits and recall on the four single gene perturbation tasks.
    Entries are mean $\pm$ sample SD of cumulative hits across five random
    seeds, with mean recall in parentheses. The second highest mean in each column is underlined.}
\label{tab:single-task-final-hits}\\
\toprule
        Method & IFN-$\gamma$ & IL-2 & NK & Carnevale \\
        \midrule
\endfirsthead
\multicolumn{5}{l}{\textit{Table~\thetable\ (continued)}}\\
\toprule
        Method & IFN-$\gamma$ & IL-2 & NK & Carnevale \\
        \midrule
\endhead
\bottomrule
\endfoot
TDAL & \benchmarkcell{$270.6\pm6.1$}{$(31.00\%)$} & \benchmarkcell{$367.4\pm3.0$}{$(42.08\%)$} & \benchmarkcell{$\underline{109.4\pm9.3}$}{$\underline{(12.49\%)}$} & \benchmarkcell{$161.2\pm8.6$}{$(18.40\%)$} \\
        \addlinespace[4pt]
        PoH & \benchmarkcell{$256.6\pm14.6$}{$(29.39\%)$} & \benchmarkcell{$\underline{347.8\pm6.0}$}{$\underline{(39.84\%)}$} & \benchmarkcell{$101.0\pm4.1$}{$(11.53\%)$} & \benchmarkcell{$145.6\pm2.7$}{$(16.62\%)$} \\
        \addlinespace[4pt]
        OAE & \benchmarkcell{$194.0\pm13.0$}{$(22.22\%)$} & \benchmarkcell{$334.2\pm6.4$}{$(38.28\%)$} & \benchmarkcell{$97.2\pm7.2$}{$(11.10\%)$} & \benchmarkcell{$124.6\pm5.8$}{$(14.22\%)$} \\
        \addlinespace[4pt]
        DiscoBAX & \benchmarkcell{$218.0\pm10.4$}{$(24.97\%)$} & \benchmarkcell{$266.0\pm17.7$}{$(30.47\%)$} & \benchmarkcell{$81.0\pm5.2$}{$(9.25\%)$} & \benchmarkcell{$151.8\pm5.9$}{$(17.33\%)$} \\
        \addlinespace[4pt]
        UCB & \benchmarkcell{$196.4\pm12.4$}{$(22.50\%)$} & \benchmarkcell{$331.0\pm10.2$}{$(37.92\%)$} & \benchmarkcell{$85.4\pm6.2$}{$(9.75\%)$} & \benchmarkcell{$123.8\pm8.3$}{$(14.13\%)$} \\
        \addlinespace[4pt]
        Expected Improvement & \benchmarkcell{$185.8\pm16.9$}{$(21.28\%)$} & \benchmarkcell{$315.4\pm7.1$}{$(36.13\%)$} & \benchmarkcell{$92.6\pm4.1$}{$(10.57\%)$} & \benchmarkcell{$123.2\pm8.3$}{$(14.06\%)$} \\
        \addlinespace[4pt]
        Thompson Sampling & \benchmarkcell{$178.2\pm12.1$}{$(20.41\%)$} & \benchmarkcell{$299.8\pm6.5$}{$(34.34\%)$} & \benchmarkcell{$98.4\pm8.4$}{$(11.23\%)$} & \benchmarkcell{$114.0\pm10.8$}{$(13.01\%)$} \\
        \addlinespace[4pt]
        TopUncertain & \benchmarkcell{$200.8\pm10.1$}{$(23.00\%)$} & \benchmarkcell{$298.6\pm13.6$}{$(34.20\%)$} & \benchmarkcell{$91.4\pm4.4$}{$(10.43\%)$} & \benchmarkcell{$127.2\pm5.5$}{$(14.52\%)$} \\
        \addlinespace[4pt]
        BADGE & \benchmarkcell{$190.8\pm8.6$}{$(21.86\%)$} & \benchmarkcell{$237.0\pm12.0$}{$(27.15\%)$} & \benchmarkcell{$99.0\pm5.2$}{$(11.30\%)$} & \benchmarkcell{$140.2\pm7.8$}{$(16.00\%)$} \\
        \addlinespace[4pt]
        CoreSet & \benchmarkcell{$\underline{259.8\pm2.2}$}{$\underline{(29.76\%)}$} & \benchmarkcell{$298.4\pm4.8$}{$(34.18\%)$} & \benchmarkcell{$114.0\pm6.1$}{$(13.01\%)$} & \benchmarkcell{$\underline{159.4\pm7.3}$}{$\underline{(18.20\%)}$} \\
        \addlinespace[4pt]
        KMeans-Embedding & \benchmarkcell{$153.2\pm11.2$}{$(17.55\%)$} & \benchmarkcell{$202.4\pm6.5$}{$(23.18\%)$} & \benchmarkcell{$82.0\pm3.4$}{$(9.36\%)$} & \benchmarkcell{$117.2\pm8.9$}{$(13.38\%)$} \\
        \addlinespace[4pt]
        Random & \benchmarkcell{$94.0\pm6.0$}{$(10.77\%)$} & \benchmarkcell{$102.2\pm7.0$}{$(11.71\%)$} & \benchmarkcell{$102.6\pm10.5$}{$(11.71\%)$} & \benchmarkcell{$103.4\pm10.8$}{$(11.80\%)$} \\
        \addlinespace[4pt]

\end{longtable}
\endgroup

\Needspace{4\baselineskip}
\subsection{Combinatorial Perturbation Results}

Table~\ref{tab:combo-task-final-hits} reports the complete endpoint results
for the four combinatorial perturbation tasks in
Figure~\ref{fig:discovery-performance}e--h, providing cumulative
hit counts for each method, variation across seeds, and recall for the discovery performance
comparison in the main text.

\begingroup
\small
\setlength{\tabcolsep}{3pt}
    \renewcommand{\arraystretch}{1.10}
\setlength{\LTleft}{\fill}
\setlength{\LTright}{\fill}
\setlength{\LTcapwidth}{\linewidth}
\begin{longtable}{@{}lcccc@{}}
\caption{Final cumulative hits and recall on the four main combinatorial perturbation tasks.
    Entries are mean $\pm$ sample SD of cumulative hits across five random
    seeds, with mean recall in parentheses. The second highest mean in each column is underlined.}
\label{tab:combo-task-final-hits}\\
\toprule
        Method & \shortstack{Simpson\\K562} & SCHEMATIC & \shortstack{Kuzmin\\yeast} & \shortstack{Horlbeck\\Jurkat} \\
        \midrule
\endfirsthead
\multicolumn{5}{l}{\textit{Table~\thetable\ (continued)}}\\
\toprule
        Method & \shortstack{Simpson\\K562} & SCHEMATIC & \shortstack{Kuzmin\\yeast} & \shortstack{Horlbeck\\Jurkat} \\
        \midrule
\endhead
\bottomrule
\endfoot
TDAL & \benchmarkcell{$\underline{6765.4\pm102.5}$}{$\underline{(94.40\%)}$} & \benchmarkcell{$160.4\pm14.2$}{$(29.01\%)$} & \benchmarkcell{$4141.0\pm194.0$}{$(38.59\%)$} & \benchmarkcell{$\underline{4542.8\pm4.8}$}{$\underline{(99.45\%)}$} \\
        \addlinespace[4pt]
        PoH & \benchmarkcell{$6771.0\pm75.7$}{$(94.47\%)$} & \benchmarkcell{$187.2\pm5.1$}{$(33.85\%)$} & \benchmarkcell{$\underline{4111.4\pm42.2}$}{$\underline{(38.31\%)}$} & \benchmarkcell{$4551.4\pm1.5$}{$(99.64\%)$} \\
        \addlinespace[4pt]
        OAE & \benchmarkcell{$5951.2\pm36.4$}{$(83.04\%)$} & \benchmarkcell{$104.8\pm10.0$}{$(18.95\%)$} & \benchmarkcell{$3420.4\pm83.1$}{$(31.87\%)$} & \benchmarkcell{$4489.2\pm23.3$}{$(98.27\%)$} \\
        \addlinespace[4pt]
        UCB & \benchmarkcell{$5955.4\pm46.0$}{$(83.09\%)$} & \benchmarkcell{$105.0\pm6.6$}{$(18.99\%)$} & \benchmarkcell{$3636.6\pm157.1$}{$(33.89\%)$} & \benchmarkcell{$4394.0\pm103.0$}{$(96.19\%)$} \\
        \addlinespace[4pt]
        Expected Improvement & \benchmarkcell{$5005.2\pm176.9$}{$(69.84\%)$} & \benchmarkcell{$117.0\pm2.9$}{$(21.16\%)$} & \benchmarkcell{$3013.4\pm278.6$}{$(28.08\%)$} & \benchmarkcell{$4134.2\pm92.5$}{$(90.50\%)$} \\
        \addlinespace[4pt]
        Thompson Sampling & \benchmarkcell{$5903.8\pm17.1$}{$(82.37\%)$} & \benchmarkcell{$99.6\pm6.9$}{$(18.01\%)$} & \benchmarkcell{$3196.2\pm164.5$}{$(29.78\%)$} & \benchmarkcell{$4044.8\pm48.9$}{$(88.55\%)$} \\
        \addlinespace[4pt]
        TopUncertain & \benchmarkcell{$3758.0\pm181.5$}{$(52.43\%)$} & \benchmarkcell{$107.8\pm7.6$}{$(19.49\%)$} & \benchmarkcell{$3759.6\pm145.6$}{$(35.03\%)$} & \benchmarkcell{$4059.8\pm136.6$}{$(88.87\%)$} \\
        \addlinespace[4pt]
        BADGE & \benchmarkcell{$932.8\pm13.3$}{$(13.02\%)$} & \benchmarkcell{$151.4\pm11.2$}{$(27.38\%)$} & \benchmarkcell{$2671.8\pm25.2$}{$(24.90\%)$} & \benchmarkcell{$1622.6\pm30.0$}{$(35.52\%)$} \\
        \addlinespace[4pt]
        CoreSet & \benchmarkcell{$1693.6\pm44.4$}{$(23.63\%)$} & \benchmarkcell{$\underline{182.0\pm8.0}$}{$\underline{(32.91\%)}$} & \benchmarkcell{$3417.0\pm101.9$}{$(31.84\%)$} & \benchmarkcell{$2845.6\pm78.3$}{$(62.29\%)$} \\
        \addlinespace[4pt]
        KMeans-Embedding & \benchmarkcell{$767.4\pm18.8$}{$(10.71\%)$} & \benchmarkcell{$149.8\pm7.8$}{$(27.09\%)$} & \benchmarkcell{$2646.6\pm49.2$}{$(24.66\%)$} & \benchmarkcell{$1316.0\pm17.8$}{$(28.81\%)$} \\
        \addlinespace[4pt]
        Random & \benchmarkcell{$1313.4\pm59.8$}{$(18.33\%)$} & \benchmarkcell{$99.2\pm10.0$}{$(17.94\%)$} & \benchmarkcell{$2603.2\pm45.3$}{$(24.26\%)$} & \benchmarkcell{$1339.0\pm25.3$}{$(29.31\%)$} \\
        \addlinespace[4pt]

\end{longtable}
\endgroup

\Needspace{4\baselineskip}
\subsection{Rank Consistency Across Tasks}
\label{app:rank-consistency}

We rank the eleven common methods by their mean final discovery counts over
five seeds on each task. DiscoBAX is evaluated on four tasks and is excluded
from this comparison. Table~\ref{tab:rank-consistency} summarizes mean ranks,
variation across tasks, and top three coverage. TDAL has the best mean rank
of 1.63 and the smallest rank standard deviation of 0.74, and is the only
method ranked among the top three on all eight tasks. Omitting each task in
turn leaves TDAL with the best average rank over the remaining seven tasks.
Thus, its leading average rank does not depend on any single task.

To assess variation across seeds, we first rank methods separately for
each task and seed. We then calculate the rank standard deviation across
the five seeds within each task and average it over tasks. TDAL has the
smallest rank variation across tasks, while CoreSet has the smallest
variation across seeds.

\begin{table}[!htb]
    \centering
    \small
    \caption{Rank consistency across the eight main tasks. Task ranks are based
    on mean final discovery counts over five seeds among the eleven common methods.
    Ties receive average ranks. Rank SD is the sample standard deviation across
    task ranks. Seed rank SD is the mean of the sample standard deviations
    across five seeds within each task.}
    \label{tab:rank-consistency}
    \setlength{\tabcolsep}{4pt}
    \renewcommand{\arraystretch}{1.15}
    \begin{tabular*}{\linewidth}{@{\extracolsep{\fill}}lccccc@{}}
        \toprule
        Method & Mean rank & Rank SD & Range & Top three & Seed rank SD \\
        \midrule
        TDAL & 1.63 & 0.74 & 1--3 & 8/8 & 0.713 \\
        PoH & 2.13 & 1.13 & 1--4 & 7/8 & 0.686 \\
        CoreSet & 4.63 & 3.16 & 1--8 & 4/8 & 0.498 \\
        OAE & 5.38 & 2.07 & 3--9 & 2/8 & 1.233 \\
        UCB & 5.63 & 2.45 & 3--10 & 1/8 & 1.131 \\
        TopUncertain & 6.00 & 1.93 & 3--9 & 1/8 & 1.004 \\
        EI & 6.75 & 1.39 & 5--8 & 0/8 & 0.878 \\
        BADGE & 7.13 & 2.47 & 4--10 & 0/8 & 0.757 \\
        TS & 7.50 & 1.93 & 5--10 & 0/8 & 1.003 \\
        KMeans-Embedding & 9.63 & 2.00 & 5--11 & 0/8 & 0.594 \\
        Random & 9.63 & 2.77 & 3--11 & 1/8 & 0.753 \\
        \bottomrule
    \end{tabular*}
\end{table}

\section{Landscape Structure and Active Discovery}
\label{app:landscape}
\renewcommand{\thetable}{\thesection\arabic{table}}
\renewcommand{\theHtable}{\thesection.\arabic{table}}
\setcounter{table}{0}
\renewcommand{\thefigure}{\thesection\arabic{figure}}
\renewcommand{\theHfigure}{\thesection.\arabic{figure}}
\setcounter{figure}{0}

\Needspace{4\baselineskip}
\subsection{Local Hit Enrichment and Discovery Gains on Real Tasks}
\label{app:real-landscape}

\textbf{Local hit enrichment.}
We compare the hit fraction around hit candidates with the overall hit
fraction in the analysis sample to measure how hits cluster in candidate
representation space. Neighbors are determined by cosine distance after
standardizing the fixed candidate representations $\phi(x)$ and applying L2 normalization.

For each task, we randomly sample hit candidates and other candidates in their proportions in
the full candidate pool, forming an analysis sample $\mathcal P$ of at most
10,000 candidates. Let $\mathcal P_+=\{x\in\mathcal P:z(x)=1\}$ be its hit set.
For each hit candidate $x$, let $\mathcal N_{10}(x)$ contain the ten nearest
other candidates within $\mathcal P$. We compute the hit fraction among these
ten neighbors, average it over all hit candidates, and divide by the overall
hit fraction in the sample to obtain
\begin{equation}
\operatorname{Enrich}_{10}
=\frac{\displaystyle
\frac{1}{|\mathcal P_+|}\sum_{x\in\mathcal P_+}
\frac{1}{10}\sum_{y\in\mathcal N_{10}(x)}z(y)}
{\displaystyle |\mathcal P_+|/|\mathcal P|}.
\label{eq:local-hit-enrichment}
\end{equation}
The numerator is the mean neighbor hit fraction around hit candidates,
and the denominator is the overall hit fraction in the analysis sample.
For example, an overall hit fraction of 5\% and a mean neighbor hit fraction
of 15\% give $3\times$ enrichment. Values above one indicate that other hits
are more common near hit candidates. Values near one indicate a similar hit
fraction to the sample overall, and values below one indicate a lower local
hit fraction. We repeat sampling and calculation three times for each task.
Table~\ref{tab:real-landscape-gains} reports the mean enrichment over these
three samples.

\textbf{Discovery gains.}
For each task, we divide each method's final cumulative hit count by Random's
count under the same seed, then average over five seeds. The shared methods
are PoH, OAE, UCB, EI, TS, TopUncertain, BADGE, CoreSet, and KMeans-Embedding.
Their average gain is the mean of these nine methods'
results. Across the eight tasks, this average gain increases with local hit
enrichment (Spearman $\rho=0.95$, $n=8$). Both enrichment and discovery gains
are near random levels on NK and higher on Simpson K562, showing a close
association between local hit structure and active discovery gains across
tasks.

\begin{table}[!htb]
    \centering
    \small
    \caption{Hit clustering and active discovery gains on real tasks.}
    \label{tab:real-landscape-gains}
    \setlength{\tabcolsep}{4pt}
    \renewcommand{\arraystretch}{1.12}
    \begin{tabular*}{\linewidth}{@{\extracolsep{\fill}}lccc@{}}
        \toprule
        Dataset & \shortstack{Hit enrichment\\(10 nearest neighbors)} & \shortstack{Shared methods\\/ Random} & TDAL / Random \\
        \midrule
        NK & $0.83\times$ & $0.94\times$ & $1.08\times$ \\
        Carnevale & $1.25\times$ & $1.27\times$ & $1.57\times$ \\
        SCHEMATIC & $1.51\times$ & $1.36\times$ & $1.63\times$ \\
        Kuzmin & $1.76\times$ & $1.28\times$ & $1.59\times$ \\
        IFN-$\gamma$ & $2.53\times$ & $2.15\times$ & $2.89\times$ \\
        IL-2 & $3.58\times$ & $2.91\times$ & $3.61\times$ \\
        Horlbeck Jurkat & $5.84\times$ & $2.61\times$ & $3.39\times$ \\
        Simpson K562 & $6.27\times$ & $3.11\times$ & $5.16\times$ \\
        \bottomrule
    \end{tabular*}
\end{table}

\Needspace{4\baselineskip}
\subsection{Noisy Multicluster Landscapes}
\label{app:multicluster}

\textbf{Landscape construction.}
We construct landscapes in which hits are dispersed or concentrated in
six clusters. The 17,660 candidates have fixed two-dimensional coordinates
that are defined without phenotype information. KMeans divides these
coordinates into six regions, each with a response peak at its center.

Within each region, candidates closer to the center receive higher response
scores. We sort candidates by distance, negate their ranks, and divide by
the region's hit quota. The quotas are 148 for one region and 147 for each
of the other five, giving 883 hits in total. This scaling produces clusters
of similar hit counts when no noise is added. We standardize the scores
across candidates and mix them with independent standardized Gaussian noise.
Increasing the noise contribution disperses hits away from the cluster centers.

To keep phenotype values comparable, we rank the mixed scores and assign
them the same fixed set of standard normal quantiles. The top 5\% are designated as
hits. All settings therefore share the phenotype distribution and 883 hits,
while hit locations range from nearly random at $1.05\times$ mean enrichment
to six concentrated clusters at $17.71\times$. Surrogates receive only
candidate coordinates and measured phenotypes.

\textbf{Screening settings.}
We use $n_0=128$, $B=32$, and $T=60$, giving a total query budget of 2,048.
Each landscape setting has five paired repetitions with independent noise and
selection seeds 5--9. Methods share the landscape, initial set, and budget
within each repetition. The $17.71\times$ landscape contains no noise, so
only selection seeds vary in this setting. Model and training settings follow the real tasks.
TDAL and PoH retain their empirical hit frequency and Gaussian tail estimators,
respectively.

\begin{figure}[!htb]
\centering
\includegraphics[width=\linewidth]{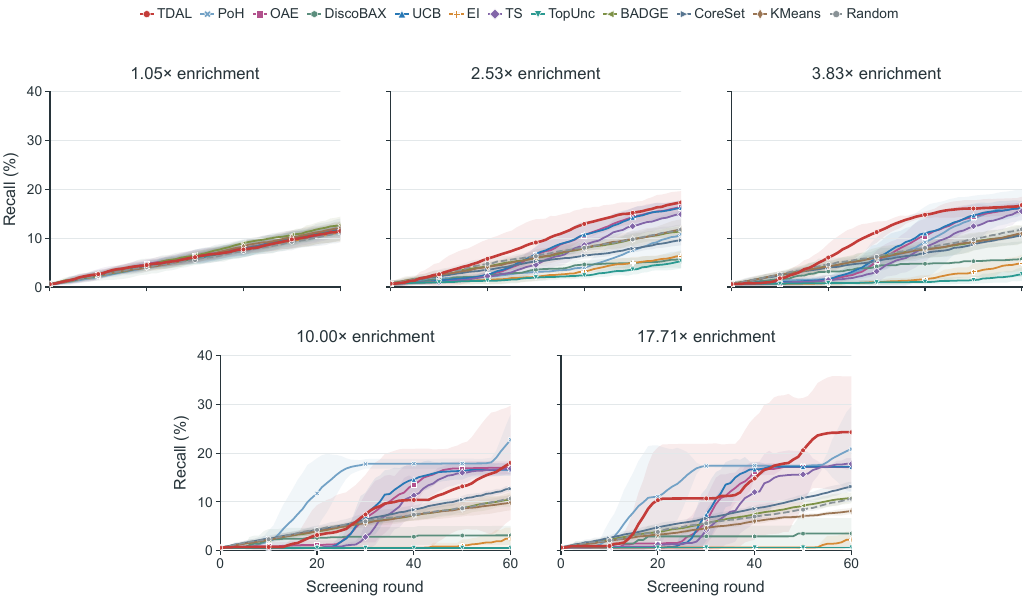}
\caption{Discovery trajectories for all twelve methods in
Figure~\ref{fig:landscape-structure}b. Panel titles give mean hit enrichment.
Lines show mean recall over five paired repetitions, with shaded 95\%
confidence intervals. Round 0 is the initial set of 128 candidates.
All panels share axis ranges.}
\label{fig:multicluster-trajectories}
\end{figure}

The comparison includes TDAL and the eleven baselines in
Figure~\ref{fig:landscape-structure}, giving 300 runs across five settings.
Their mean hit enrichments are $1.05$, $2.53$, $3.83$, $10.00$,
and $17.71\times$. Enrichment is computed using ten nearest neighbors in
the full pool with Euclidean distances in the fixed coordinates, then
averaged over the five realizations. The four snapshots in
Figure~\ref{fig:landscape-structure}b show the first realization with its own
enrichment. Table~\ref{tab:multicluster-results} reports final recall for
these same methods and settings, and Figure~\ref{fig:multicluster-trajectories}
shows their discovery trajectories.

\textbf{Discovery performance.}
With nearly random hit locations, mean discovery gains are close to one.
Between the least and most clustered settings, TDAL's mean gain over
Random rises from $1.04\times$ to $2.26\times$, and PoH's from
$1.06\times$ to $1.94\times$. OAE, UCB, and TS reach gains of
$1.62\times$, $1.62\times$, and $1.67\times$, respectively, at the highest
enrichment. Thus, clustered hits provide useful information for several
acquisition functions, although their gains differ across settings.
BADGE and KMeans-Embedding do not improve consistently with clustering, while EI,
TopUncertain, and DiscoBAX remain below Random in the four structured settings.

\begin{table}[!htb]
\centering
\fontsize{8}{10}\selectfont
\caption{Final recall (\%) on the five multicluster landscapes in
Figure~\ref{fig:landscape-structure}b. Column headings give mean hit enrichment.
Values are means $\pm$ sample standard deviations over five paired repetitions.}
\label{tab:multicluster-results}
\setlength{\tabcolsep}{3pt}
\renewcommand{\arraystretch}{1.18}
\begin{tabular*}{\linewidth}{@{\extracolsep{\fill}}lccccc@{}}
\toprule
Method & $1.05\times$ & $2.53\times$ & $3.83\times$ & $10.00\times$ & $17.71\times$ \\
\midrule
\textbf{TDAL} & $11.48\pm0.39$ & $17.33\pm1.87$ & $16.78\pm1.38$ & $18.01\pm9.40$ & $24.28\pm9.23$ \\
PoH & $11.66\pm1.05$ & $10.83\pm4.00$ & $16.76\pm2.62$ & $22.74\pm4.19$ & $20.82\pm7.05$ \\
OAE & $11.78\pm0.60$ & $16.49\pm1.11$ & $16.24\pm0.74$ & $16.99\pm1.07$ & $17.17\pm0.17$ \\
DiscoBAX & $11.51\pm1.56$ & $5.73\pm1.46$ & $5.73\pm2.92$ & $3.15\pm1.57$ & $3.53\pm2.52$ \\
UCB & $11.73\pm0.75$ & $16.19\pm1.30$ & $16.26\pm1.09$ & $16.60\pm1.01$ & $17.19\pm0.17$ \\
EI & $11.33\pm1.62$ & $6.34\pm0.76$ & $4.92\pm1.37$ & $2.45\pm1.85$ & $2.27\pm1.35$ \\
TS & $11.39\pm1.08$ & $14.93\pm1.82$ & $15.52\pm1.41$ & $16.81\pm1.00$ & $17.80\pm0.95$ \\
TopUncertain & $11.71\pm0.98$ & $5.30\pm1.21$ & $2.63\pm1.08$ & $0.61\pm0.10$ & $0.66\pm0.17$ \\
BADGE & $12.75\pm1.33$ & $11.42\pm0.83$ & $10.80\pm1.47$ & $10.51\pm1.19$ & $10.83\pm0.87$ \\
CoreSet & $11.60\pm1.95$ & $9.60\pm0.85$ & $10.55\pm1.25$ & $12.71\pm0.43$ & $13.16\pm0.78$ \\
KMeans-Embedding & $12.07\pm0.94$ & $11.87\pm1.60$ & $11.01\pm0.67$ & $9.90\pm1.31$ & $8.13\pm0.49$ \\
Random & $11.17\pm1.30$ & $11.80\pm0.79$ & $11.82\pm0.55$ & $10.69\pm0.57$ & $10.65\pm0.45$ \\
\bottomrule
\end{tabular*}
\end{table}

\section{Ablation Studies and Selection Dynamics}
\label{app:ablations-dynamics}
\renewcommand{\thetable}{\thesection\arabic{table}}
\renewcommand{\theHtable}{\thesection.\arabic{table}}
\renewcommand{\thefigure}{\thesection\arabic{figure}}
\renewcommand{\theHfigure}{\thesection.\arabic{figure}}
\setcounter{table}{0}
\setcounter{figure}{0}

\subsection{Component Ablations}
\label{app:component-ablations}

We compare five ablations with full TDAL on IFN-$\gamma$, IL-2, Kuzmin, and
Horlbeck Jurkat. All methods share the surrogate configuration, fixed
candidate representations, initial sets, experimental budgets, and five
random seeds.
Final cumulative hit counts are reported in Table~\ref{tab:component-ablations}.

\textbf{Surrogate only.}
This ablation refits the same surrogate as TDAL after each round of
observations but selects candidates with the highest $\hat{p}_t(x)$
without training an acquisition network, breaking exact probability ties at random.
Its mean final recall is 0.54--13.01 percentage points lower than full TDAL across the four tasks.
Updating response predictions alone therefore does not recover the full
model's discovery gains. Learning a selection function from batch feedback
further improves screening.

\textbf{w/o adaptive $w_t$.}
This ablation retains successor value bootstrapping but replaces the
adaptive Hellinger weight with 1, applying the same successor
weight in every round. Mean final recall decreases by 3.88--9.03 percentage
points across the four tasks. Giving successor value full weight at every round therefore performs
less well than adjusting its contribution to the observed prediction changes.

\textbf{w/o LBO.}
This ablation disables LBO transitions constructed from the initial set,
the associated initial TD training, and online LBO. The acquisition network
learns only after receiving real batch feedback. Mean final recall is
0.42--5.54 percentage points lower than full TDAL across the four tasks.
LBO reuses measured outcomes to provide training experience at
initialization and supplement the single real transition obtained per
online round.

\textbf{w/o successor.}
This ablation sets the training target to $Y=R$, retaining the
hit probability term, acquisition network, and both real and LBO transitions.
With surrogate predictions held fixed, the network fits the difference
between the observed batch discovery count and its surrogate estimate.
Mean final recall exceeds Surrogate only on all four tasks but is
0.15--1.21 percentage points lower than full TDAL. Current feedback can
therefore train an effective selection function, while successor value
bootstrapping further increases final discoveries.

\textbf{w/o anchor.}
This ablation uses $q_\theta(s_t,x)=f_\theta(s_t,c_t(x))$.
Both current and successor batch values sum network outputs without
explicitly adding $\hat{p}_t(x)$. Surrogate prediction features remain
in the inputs. Relative to full TDAL, mean final recall decreases by
0.96 and 2.82 percentage points on IFN-$\gamma$ and IL-2, respectively,
and increases by 0.31 and 0.16 points on Kuzmin and Horlbeck Jurkat.
The two single perturbation tasks each use 60 screening rounds, whereas
the two combinatorial tasks each use 200. More rounds provide the acquisition
network with more feedback and training opportunities. The explicit
hit probability term is more beneficial on the two tasks with 60 rounds.
This pattern is consistent with a useful probability reference early in
learning and motivates gradually reducing its contribution as feedback
accumulates. Testing such a schedule within the same task would separate
the effect of screening horizon from differences between datasets.

\subsection{Training Dynamics and Candidate Reranking}
\label{app:training-dynamics}

\textbf{Candidates analyzed.}
On IL-2, 85 hit candidates are discovered during active screening in all
five TDAL runs and none of the five Surrogate only runs. We compare their
discovery times and selection ranks across the five random seeds.

\textbf{Training and discovery across seeds.}
Figure~\ref{fig:il2-multiseed-dynamics}a--b shows TD loss per round and
cumulative discoveries among these 85 candidates. Early loss changes
vary in direction and magnitude across seeds. For all five seeds,
fluctuations are smaller in the last 30 than in the first 10 rounds.
Discoveries concentrate in the second half, with 74.1\%--87.1\% of the set
discovered in the last 30 rounds (Table~\ref{tab:il2-multiseed-selection}).
This common timing pattern shows that these candidates mostly enter
selected batches after feedback accumulates.

\textbf{Candidate reranking at selection.}
We compare each candidate's TDAL score rank and concurrent surrogate
hit probability rank at selection, using the same unmeasured pool and
surrogate predictions. Across seeds, mean TDAL ranks are 16.1--17.2,
whereas mean surrogate ranks are 9,442--9,780. Of 425 selection records,
413 have surrogate ranks outside the top $B=32$ (97.2\%,
Figure~\ref{fig:il2-multiseed-dynamics}c--g). Across seeds, the acquisition
network consistently promotes candidates ranked low by concurrent
surrogate predictions into the selected batch.

\begingroup
\setlength{\intextsep}{8pt}
\begin{figure}[!htb]
    \centering
    \includegraphics[width=0.90\linewidth]{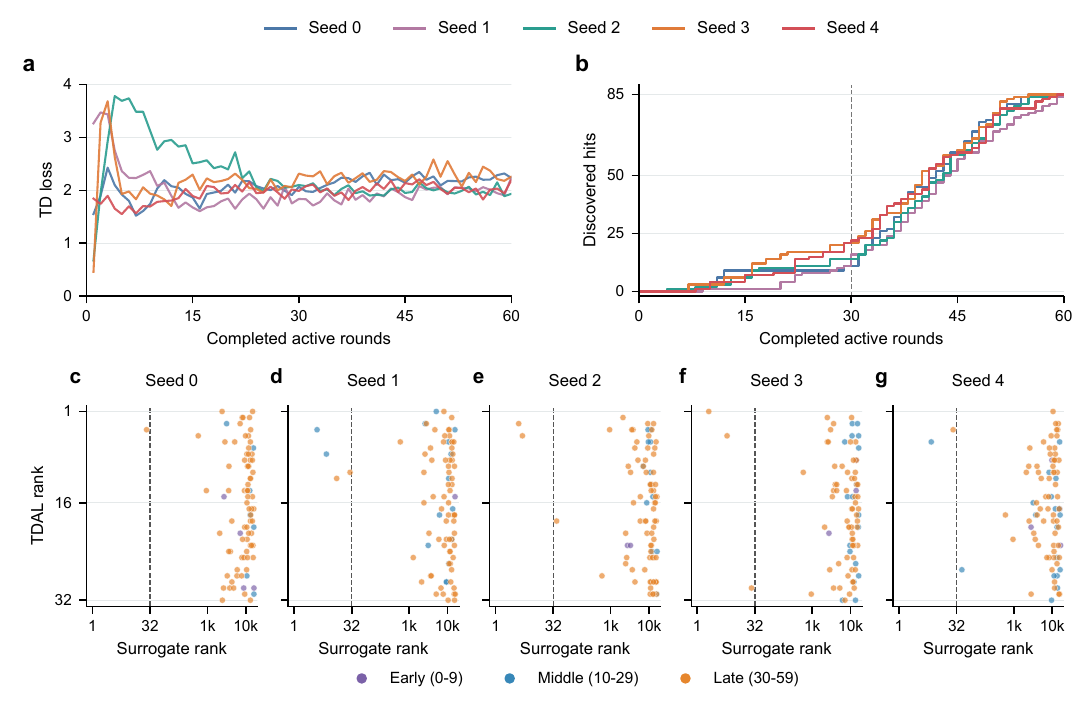}
    \caption{Training dynamics and candidate reranking on IL-2 across five
    random seeds.
    \textbf{a,} TD loss per round.
    \textbf{b,} Cumulative discoveries among the 85 hit candidates analyzed.
    The dashed line marks 30 completed rounds.
    \textbf{c--g,} Surrogate and TDAL ranks at selection for seeds 0--4,
    with a logarithmic horizontal axis and a dashed line at surrogate
    rank 32. Colors indicate selection in rounds 0--9, 10--29, and 30--59.}
    \label{fig:il2-multiseed-dynamics}
\end{figure}

\begin{table}[!htb]
    \centering
    \small
    \caption{Discovery stages and selection ranks of the hit candidates
    analyzed across five seeds. Stage columns count newly discovered
    candidates. Ranks are computed in the same unmeasured pool at selection
    and averaged over the 85 candidates.}
    \label{tab:il2-multiseed-selection}
    \setlength{\tabcolsep}{4pt}
    \renewcommand{\arraystretch}{1.12}
    \begin{tabular*}{\linewidth}{@{\extracolsep{\fill}}lccccc@{}}
        \toprule
        Seed & \shortstack{Rounds\\0--9} & \shortstack{Rounds\\10--29}
        & \shortstack{Rounds\\30--59} & \shortstack{Mean TDAL\\rank}
        & \shortstack{Mean surrogate\\rank} \\
        \midrule
        0 & 4 & 7 & 74 & 16.6 & 9,780 \\
        1 & 1 & 15 & 69 & 16.1 & 9,442 \\
        2 & 2 & 12 & 71 & 16.8 & 9,574 \\
        3 & 3 & 18 & 64 & 17.2 & 9,522 \\
        4 & 4 & 18 & 63 & 16.4 & 9,479 \\
        \bottomrule
    \end{tabular*}
\end{table}
\endgroup

% F.3 is hidden from the compiled manuscript. Remove this conditional to restore it.
\iffalse
\Needspace{4\baselineskip}
\subsection{Fixed and Adaptive Feedback Weights}
\label{app:fixed-weights}

We compare the adaptive Hellinger weight with eleven constant successor
weights $w\in\{0,0.1,\ldots,1\}$ on IFN-$\gamma$ and IL-2. Each setting uses
five seeds, $n_0=128$, $B=32$, and 60 screening rounds, with the same
surrogate, acquisition network, LBO, initial sets, and budget.
Only the successor weight changes. The settings $w=0$ and $w=1$ correspond to the
w/o successor and w/o adaptive $w_t$ ablations.

Figure~\ref{fig:weight-trajectories} shows the corresponding discovery trajectories.

\begin{figure}[!htb]
    \centering
    \includegraphics[width=\linewidth]{figures/appendix_F3.pdf}
    \caption{Discovery trajectories for fixed and adaptive successor
    weights on \textbf{a,} IFN-$\gamma$ and \textbf{b,} IL-2. Lines show
    mean recall across five seeds, with shaded 95\% confidence intervals.
    Round 0 corresponds to the initial set of 128 measured candidates.}
    \label{fig:weight-trajectories}
\end{figure}
\fi

\Needspace{4\baselineskip}
\section{Batch Size Sensitivity and Computational Cost}
\label{app:batch-efficiency}
\renewcommand{\thetable}{\thesection\arabic{table}}
\renewcommand{\theHtable}{\thesection.\arabic{table}}
\renewcommand{\thefigure}{\thesection\arabic{figure}}
\renewcommand{\theHfigure}{\thesection.\arabic{figure}}
\setcounter{table}{0}
\setcounter{figure}{0}

\Needspace{4\baselineskip}
\subsection{Batch Size Sensitivity}
\label{app:batch-sensitivity}

\textbf{Experimental settings.}
We vary batch size at fixed initial sample size and query budget on IL-2
and Simpson K562 to compare discovery at different feedback frequencies.
IL-2 uses $n_0=128$, $M=2{,}048$, and $B\in\{16,32,64,128\}$.
Simpson K562 uses $n_0=1{,}024$, $M=26{,}624$, and
$B\in\{64,128,256,512\}$. Tables~\ref{tab:batch-il2}--\ref{tab:batch-simpson}
give the corresponding numbers of rounds. TDAL, PoH, UCB, BADGE, and CoreSet
retain their respective model and training hyperparameters and share the
candidate pool, initial set, and budget within each task and seed.
Each setting uses five seeds.

\textbf{Single perturbations.}
On IL-2, TDAL has the highest mean at $B=16,32,64$. At $B=128$, TDAL's mean
decreases to 252.2 with greater variation across seeds, while PoH reaches
351.4. This setting provides only 15 rounds of experimental feedback and
no initial LBO, illustrating performance under limited training experience.

\begin{table}[H]
\centering
\small
\caption{Batch size comparison at a fixed budget on IL-2. Values are final cumulative hits, reported as
mean $\pm$ sample standard deviation over five seeds. The highest mean in
each column is bold and the second highest is underlined.}
\label{tab:batch-il2}
\setlength{\tabcolsep}{3pt}
\renewcommand{\arraystretch}{1.15}
\begin{tabular*}{\linewidth}{@{\extracolsep{\fill}}lcccc@{}}
\toprule
Method & \shortstack{$B=16$\\$T=120$} & \shortstack{$B=32$\\$T=60$} & \shortstack{$B=64$\\$T=30$} & \shortstack{$B=128$\\$T=15$} \\
\midrule
\textbf{TDAL} & $\mathbf{355.4\pm7.8}$ & $\mathbf{367.4\pm3.0}$ & $\mathbf{353.8\pm10.1}$ & $252.2\pm59.5$ \\
PoH & $\underline{348.8\pm7.5}$ & $\underline{347.8\pm6.0}$ & $\underline{347.0\pm11.0}$ & $\mathbf{351.4\pm8.0}$ \\
UCB & $336.6\pm6.2$ & $331.0\pm10.2$ & $340.0\pm18.4$ & $\underline{333.2\pm15.3}$ \\
BADGE & $263.8\pm3.6$ & $237.0\pm12.0$ & $216.8\pm9.3$ & $205.8\pm8.5$ \\
CoreSet & $301.2\pm8.8$ & $298.4\pm4.8$ & $297.0\pm5.8$ & $297.2\pm9.4$ \\
\bottomrule
\end{tabular*}
\end{table}

\Needspace{4\baselineskip}
\textbf{Double perturbations.}
On Simpson K562, TDAL stays in the top two at all four batch sizes, close
to PoH. TDAL has the highest means at $B=64,256$, while PoH is slightly
higher at $B=128,512$. TDAL's mean declines with larger batches,
remaining above UCB, BADGE, and CoreSet.
At a fixed query budget, larger batches leave fewer rounds for incorporating
experimental feedback and training the acquisition function. Smaller batches
provide more opportunities to update selection, while PoH remains
competitive at larger batches.

\begin{table}[H]
\centering
\small
\caption{Batch size comparison at a fixed budget on Simpson K562. Values are final cumulative hits, reported as
mean $\pm$ sample standard deviation over five seeds. The highest mean in
each column is bold and the second highest is underlined.}
\label{tab:batch-simpson}
\setlength{\tabcolsep}{3pt}
\renewcommand{\arraystretch}{1.15}
\begin{tabular*}{\linewidth}{@{\extracolsep{\fill}}lcccc@{}}
\toprule
Method & \shortstack{$B=64$\\$T=400$} & \shortstack{$B=128$\\$T=200$} & \shortstack{$B=256$\\$T=100$} & \shortstack{$B=512$\\$T=50$} \\
\midrule
\textbf{TDAL} & $\mathbf{6868.2\pm65.8}$ & $\underline{6765.4\pm102.5}$ & $\mathbf{6669.2\pm153.1}$ & $\underline{6638.0\pm81.9}$ \\
PoH & $\underline{6775.8\pm41.5}$ & $\mathbf{6771.0\pm75.7}$ & $\underline{6655.6\pm78.8}$ & $\mathbf{6670.8\pm80.0}$ \\
UCB & $5955.6\pm52.5$ & $5955.4\pm46.0$ & $5935.4\pm50.3$ & $5874.0\pm25.6$ \\
BADGE & $921.2\pm27.9$ & $932.8\pm13.3$ & $1006.8\pm59.0$ & $1015.8\pm40.4$ \\
CoreSet & $1827.0\pm38.0$ & $1693.6\pm44.4$ & $1522.2\pm32.9$ & $1377.6\pm27.3$ \\
\bottomrule
\end{tabular*}
\end{table}

\Needspace{4\baselineskip}
\subsection{Computational Cost}
\label{app:computational-cost}

On IL-2, we use $n_0=128$, $B=32$, and the full budget $M=2{,}048$,
measuring active rounds 2--4 for each method with seeds 0--2. Each run has
exclusive access to one NVIDIA A800 80 GB GPU and eight CPU cores, with runs for different seeds executed sequentially. Mean time per round excludes initialization
and the first round, and includes selection and all model updates within
the measured rounds. Peak host memory and reserved GPU memory are recorded
over the same window, including persistent model and replay state.

\begin{figure}[!htb]
    \centering
    \includegraphics[width=0.90\linewidth]{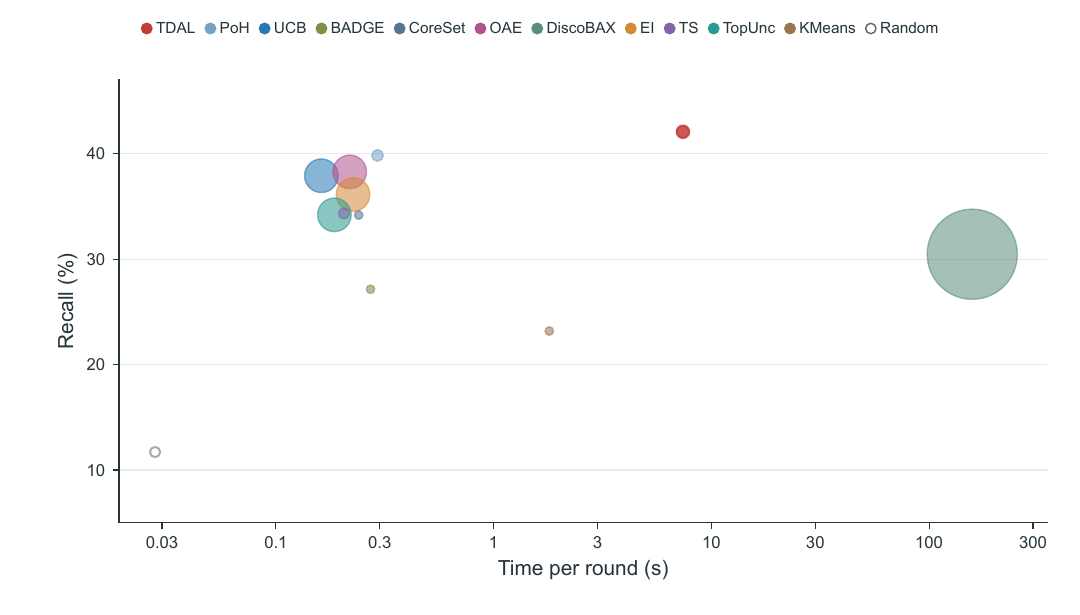}
    \caption{Discovery performance and computational cost on IL-2.
    The horizontal axis shows mean time per round on a logarithmic scale,
    and the vertical axis shows final hit recall for twelve methods. Circle area is proportional
    to peak reserved GPU memory. Random uses zero reserved GPU memory and
    is shown as a hollow circle. Time and memory are means over three seeds
    over rounds 2--4. Recall is the mean over five
    seeds in the full screening experiment of 60 rounds.}
    \label{fig:il2-efficiency-performance}
\end{figure}

TDAL achieves 42.08\% final recall on IL-2, with a mean time of 7.40 seconds
per round, 1.12 GiB peak host memory, and 0.051 GiB peak reserved GPU memory
(Figure~\ref{fig:il2-efficiency-performance}). DiscoBAX takes 157.78 seconds
per round, 21.3 times the TDAL runtime, and reserves 2.674 GiB of GPU memory.

\end{document}